\documentclass{article}

\PassOptionsToPackage{numbers,compress}{natbib}      
\usepackage[main, final]{neurips_2026}

\usepackage[utf8]{inputenc}
\usepackage[T1]{fontenc}

\usepackage{hyperref}
\usepackage{url}
\usepackage{graphicx}
\usepackage{booktabs}
\usepackage{amsmath}
\usepackage{amssymb}
\usepackage{amsfonts}
\usepackage{xcolor}
\usepackage{subcaption}
\usepackage{multirow}
\usepackage[capitalise,noabbrev]{cleveref}         
\usepackage{enumitem}
\usepackage{arydshln}
\usepackage{nicefrac}
\usepackage{microtype}

\newcommand{\mmethod}{\textsc{Click3R}}

\title{\mmethod{}: Interactive Stereo 3D Reconstruction with Sparse Correspondence Clicks}

\author{%
  Hao Gu$^{*\dagger}$
  \qquad Jin Yao$^{*}$
  \qquad Zezhou Cheng \\
  University of Virginia \\
\href{https://uva-computer-vision-lab.github.io/click3r}{\textcolor[RGB]{0,90,180}{\nolinkurl{uva-computer-vision-lab.github.io/click3r}}}
}

\begin{document}

\maketitle
{\renewcommand{\thefootnote}{}%
 \footnotetext{$^{*}$Equal contribution. \quad $^{\dagger}$Project lead.}}

\begin{abstract}
Despite impressive progress, existing feed-forward large 3D reconstruction models produce a single, irrevocable prediction. When these models fail, there is no mechanism to correct their mistakes.
To address this limitation, 
we introduce \mmethod{}, a framework for \emph{interactive geometric correction} of feed-forward models using sparse human correspondence clicks.
Given an image pair and a small set of cross-view point correspondences provided by a user, 
\mmethod{} injects geometric constraints into the 3D reconstruction networks through a lightweight point correspondence adapter, enabling the model to resolve ambiguities in challenging scenarios such as repetitive textures, symmetric structures, large viewpoint changes, or textureless regions.
To evaluate interactive reconstruction, we curate a challenging benchmark specifically designed to expose failure modes of existing feed-forward 3D reconstruction methods. Experiments show that \mmethod{} reduces reconstruction error with as few as a single correspondence click, improving performance on both standard and our curated benchmarks.
Our results highlight that sparse human interaction provides an effective and practical mechanism for correcting geometric reconstruction errors.

\end{abstract}

\section{Introduction}
\label{sec:intro}

\begin{figure}[t]
  \centering
  \includegraphics[width=\textwidth]{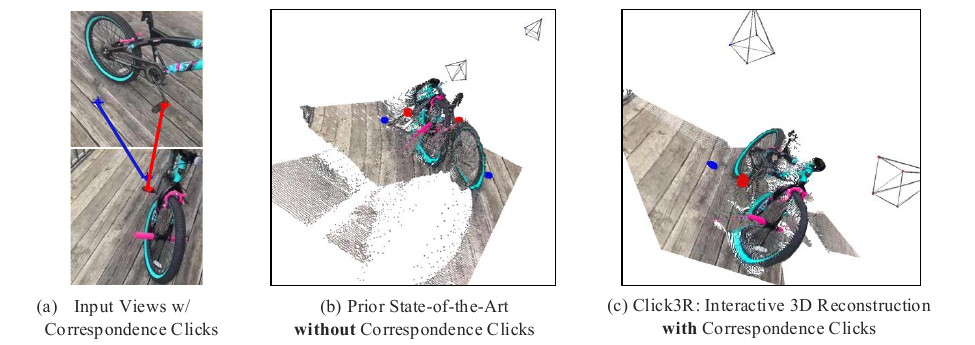}
  \caption{\textbf{Overview.} Given (a) an image pair and sparse correspondences (\textit{colored dot clicks}), \mmethod{} conditions DUSt3R~\cite{dust3r} to produce accurate 3D reconstruction. (b) DUSt3R fails on a scene with limited overlaps.
  (c) With as few as two clicks, \mmethod{} recovers correct geometry. 
  }
  \label{fig:teaser}
\end{figure}

Feed-forward stereo 3D reconstruction methods~\cite{dust3r,mast3r,rojas2025hamst3r,wen2025foundationstereo} have transformed multi-view geometry by replacing iterative optimization~\cite{mildenhall2021nerf,colmap} with direct regression.
Given two images, these models predict dense 3D pointmaps in a single forward pass, achieving remarkable speed and generalization across diverse scenes.
This paradigm has recently been extended to multi-view settings~\cite{vggt,pi3,fast3r,mvdust3rplus,must3r}.
Despite their impressive performance, the efficiency comes at a fundamental cost: \emph{when the prediction is wrong, there is no simple mechanism to correct it.}
Unlike optimization-based methods~\cite{colmap} that iteratively refine their estimates using geometric constraints, feed-forward models produce a single, irrevocable output and fail silently, particularly on challenging scenes with repetitive textures, symmetric structures, or extreme viewpoint changes (Fig.~\ref{fig:teaser}a,b).

Several approaches aim to improve the reliability of feed-forward reconstruction models through test-time optimization~\cite{lora3d,chen2025ttt3r}.
For instance, LoRA3D~\cite{lora3d} introduces a self-calibration pipeline that fine-tunes a pretrained DUSt3R model~\cite{dust3r} to a target scene.
Although these methods improve performance, iterative test-time optimization is computationally expensive and, more importantly, cannot resolve failures when the model encounters genuinely ambiguous scenes at inference time.
Another line of work improves robustness by incorporating auxiliary geometric priors.
For example, Pow3R~\cite{pow3r} shows that conditioning on camera intrinsics, relative camera poses, or depth maps can significantly improve reconstruction performance.
However, such auxiliary information is often unavailable in practice.
Meanwhile, classical reconstruction methods have long relied on explicit correspondences as geometric constraints~\cite{colmap,superglue,lightglue}.
Unfortunately, automatically estimated correspondences remain unreliable in scenarios with repetitive textures or large viewpoint changes.

Nevertheless, cross-view correspondences, even just a handful of point pairs, carry powerful geometric information: they constrain the epipolar geometry, resolve depth ambiguities, and disambiguate symmetric structures.
This motivates a simple question: \textbf{can sparse human-provided correspondences guide feed-forward 3D reconstruction models without sacrificing the efficiency of the feed-forward paradigm?}
Instead of relying solely on feed-forward models or automatically estimated correspondences, we allow the user to manually provide a small set of cross-view point correspondences (Fig.~\ref{fig:teaser}a),
when the model produces unsatisfactory reconstructions.

To unlock this capability, we introduce \textbf{\mmethod{}}, a lightweight framework that incorporates user-provided correspondences into feed-forward 3D reconstruction, enabling \textbf{interactive geometric correction without test-time model fine-tuning or optimization}.
Specifically, \mmethod{} employs a simple correspondence adapter that injects the geometric constraints imposed by user clicks into the reconstruction network via a cross-attention mechanism (Fig.~\ref{fig:architecture}).
We train the base model and the adapter jointly end-to-end on widely used 3D datasets~\cite{co3d,arkitscenes,scannetpp} using simulated human clicks. 
During training, the number of clicks is randomly sampled to expose the model to varying levels of user interaction, enabling it to operate under different levels of click sparsity at test time.

To systematically evaluate the proposed interactive reconstruction, we further introduce \textbf{Hard3DBench}, a benchmark designed to expose failure modes of state-of-the-art feed-forward reconstruction models.
The benchmark contains image pairs and 5-view tuples with ground-truth 3D reconstructions, exhibiting challenging conditions (large viewpoint changes, low overlap, large baselines, extreme depth ranges) where feed-forward reconstruction models often struggle.
Our experiments demonstrate that \mmethod{} consistently outperforms baseline methods, particularly on challenging scenarios from Hard3DBench (Fig.~\ref{fig:teaser}c).

Our contributions are summarized as follows:
\begin{itemize}[itemsep=0pt, topsep=1pt, leftmargin=15pt]
\item We propose \mmethod{}, which injects sparse human correspondence clicks into feed-forward stereo 3D reconstruction networks via a lightweight correspondence adapter.

\item We curate Hard3DBench, a diagnostic benchmark with a detailed failure taxonomy under which current feed-forward reconstruction models degrade substantially relative to standard evaluation samples.

\item We show that \mmethod{} dramatically improves reconstruction accuracy on challenging cases.
To the best of our knowledge, this is the first work on interactive stereo feed-forward 3D reconstruction using sparse correspondence clicks.
\end{itemize}

\section{Related Work}
\label{sec:related}

\paragraph{Feed-Forward Stereo 3D Reconstruction.}
DUSt3R~\cite{dust3r} pioneered dense 3D pointmap regression from image pairs using a CroCo-pretrained~\cite{croco,crocov2} ViT encoder-decoder, bypassing explicit matching and triangulation.
MASt3R~\cite{mast3r} extended this with dense local feature heads.
The paradigm has rapidly advanced toward multi-view: Spann3R~\cite{spann3r} and MUSt3R~\cite{must3r} use spatial memory across frames; CUT3R~\cite{cut3r} employs a recurrent model with persistent state; Fast3R~\cite{fast3r} scales to 1000+ images in a single forward pass; and VGGT~\cite{vggt} and Pi3~\cite{pi3} predict from arbitrary views with a single Transformer.
Despite their diversity, all these methods rely solely on learned visual priors and cannot incorporate external geometric evidence at inference.
Our work is orthogonal: \mmethod{} augments feed-forward stereo reconstruction with sparse correspondence guidance.

\paragraph{Conditioned 3D Reconstruction.}
Several concurrent methods condition feed-forward 3D models on external geometric signals: Pow3R~\cite{pow3r} on camera intrinsics, depth, or pose, G-CUT3R~\cite{gcut3r} on camera and depth modalities, Rig3R~\cite{rig3r} on multi-camera rig metadata, and Mono3R~\cite{mono3r} on monocular depth.
MapAnything~\cite{mapanything} unifies a flexible mix of conditioning inputs (intrinsics, poses, depth, partial reconstructions) into a single feed-forward model, while Any4D~\cite{any4d} extends conditional reconstruction to dynamic 4D scenes with multi-modal sensor inputs.
In the depth estimation domain, PromptDA~\cite{promptda} prompts Depth Anything~\cite{depthanything} with sparse LiDAR points.
All these methods condition on \emph{scene-level} priors that improve metric accuracy but cannot resolve perceptual aliasing.
To our knowledge, \mmethod{} is the first to introduce \emph{correspondence prompting}, which conditions feed-forward 3D reconstruction on cross-view point correspondences that directly disambiguate geometric ambiguities, analogous to SAM's~\cite{sam,sam2} click-based segmentation prompting.

\paragraph{Interactive Vision System.}
Interactive segmentation has a long history of incorporating user guidance into visual recognition.
Early methods use bounding boxes~\cite{grabcut} or scribbles~\cite{scribblesup}, while recent approaches adopt click-based prompting~\cite{dios,focalclick,simpleclick,ritm}, culminating in SAM~\cite{sam,sam2}.
Beyond segmentation, point-based interaction has been extended to image editing: DragGAN~\cite{draggan} and DragDiffusion~\cite{dragdiffusion} enable users to manipulate content by dragging points.
Recent work has also explored interactive guidance for 3D reconstruction; for example, Toon3D~\cite{weber2024toon3d} uses sparse user annotations for cartoon 3D reconstruction.
However, these interactive paradigms have largely remained confined to 2D tasks or stylized 3D settings.
Early interactive systems also explored correspondence-based 3D reconstruction~\cite{hertzmann1999interactive}.
\mmethod{} brings this idea to feed-forward stereo models, with point correspondences as geometric constraints rather than semantic cues.

\section{Methodology}
\label{sec:method}

\begin{figure}[t]
  \centering
\includegraphics[width=\textwidth]{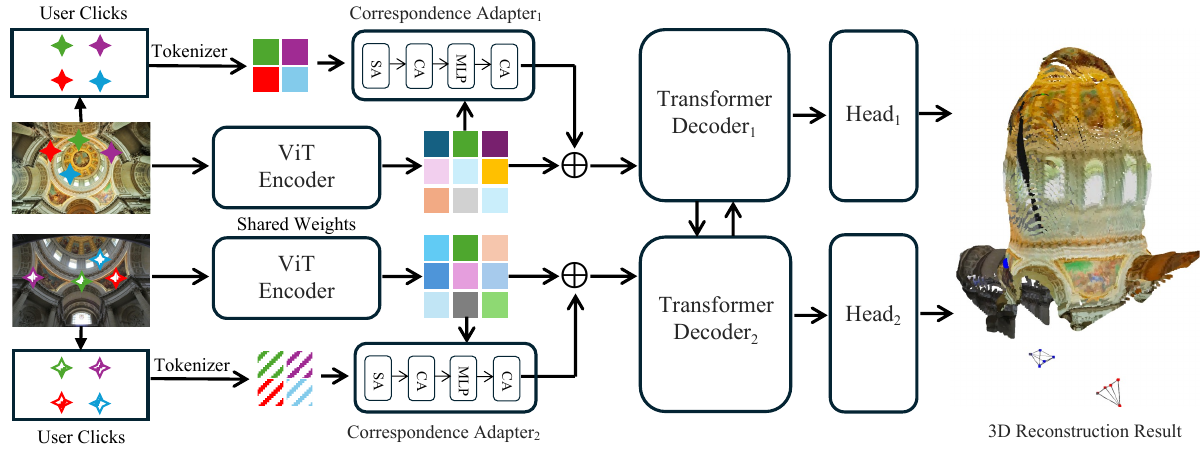}
  \caption{\textbf{\mmethod{} architecture.} A shared ViT encoder processes the two input images. Correspondence prompts are tokenized using our correspondence tokenizer (\cref{sec:prompt_encoder}). A symmetric point correspondence adapter, implemented as a lightweight transformer with self-attention and cross-attention (\cref{sec:adapter}), updates the decoder-input features for each view. The asymmetric decoder then interprets the updated embeddings and passes them to the original DUSt3R~\cite{dust3r} heads to predict 3D pointmaps and confidence. 
  }
  \label{fig:architecture}
\end{figure}

Given an image pair $(I_1, I_2)$ and an optional set of sparse point correspondences $\mathcal{C} = \{(\mathbf{p}_i^1, \mathbf{p}_i^2)\}_{i=1}^{K}$, where $\mathbf{p}_i^v \in \mathbb{R}^2$ denotes the pixel coordinates in view $v$, \mmethod{} predicts dense 3D pointmaps for both views. When no correspondences are provided, the adapter reduces to an identity mapping, and the model behaves identically to the base reconstruction model without prompts. An overview of the \mmethod{} architecture is shown in Fig.~\ref{fig:architecture}.

\subsection{Preliminaries: DUSt3R}
\label{sec:prelim}

DUSt3R~\cite{dust3r} uses a shared ViT-Large encoder ($D_\text{enc}=1024$, 24 blocks) to process each image independently, followed by an asymmetric decoder ($D_\text{dec}=768$, 12 blocks per branch) with cross-attention between views. The encoder produces patch tokens
\begin{equation}
    \mathbf{F}_v = \text{Encoder}(I_v) \in \mathbb{R}^{N \times D_\text{enc}},
    \quad N = \tfrac{H}{p} \times \tfrac{W}{p},
\end{equation}
where $p$ is the patch size. Tokens are projected to decoder dimension, $\mathbf{G}_v=\text{proj}(\mathbf{F}_v)\in\mathbb{R}^{N\times D_\text{dec}}$, then processed by interleaved self- and cross-attention blocks. A linear prediction head maps the decoded tokens to dense 3D pointmaps $\hat{X}_v \in \mathbb{R}^{H \times W \times 3}$ and confidence maps $\hat{C}_v \in \mathbb{R}^{H \times W}$ in view~1's frame.

\mmethod{} inserts its correspondence adapter after this projection and before the first decoder block, modifying $\mathbf{G}_v$ using sparse correspondence prompts. The base model and adapter are trained jointly end-to-end.

\subsection{Correspondence Tokenizer}
\label{sec:prompt_encoder}

Our tokenizer converts sparse correspondences into tokens for cross-attention with image features. For each correspondence point $\mathbf{p}_i^v=(x_i,y_i)$, we normalize its coordinates to the token grid:
\begin{equation}
    \bar{\mathbf{p}}_i^v =
    \left(\frac{x_i / p_w}{W_t - 1},\
    \frac{y_i / p_h}{H_t - 1}\right),
\end{equation}
clipped to $[0,1]$, where $p_w,p_h$ are effective patch sizes and $W_t,H_t$ are token-grid dimensions; this is the scale on which the image positional encoding places token $j$ at $j/(W_t-1)$. We encode the normalized point using SAM-style random Fourier features~\cite{sam}:
\begin{equation}
    \mathbf{e}_\text{pos}(\bar{\mathbf{p}}_i^v)
    =
    [\sin(2\pi \mathbf{B}(2\bar{\mathbf{p}}_i^v-1));\
    \cos(2\pi \mathbf{B}(2\bar{\mathbf{p}}_i^v-1))]
    \in \mathbb{R}^{D},
\end{equation}
with fixed Gaussian matrix $\mathbf{B}\in\mathbb{R}^{D/2\times2}$, preserving sub-pixel precision.

To identify the view and to link matched points across views, we add a learnable image embedding $\mathbf{e}_\text{img}(v)\in\mathbb{R}^{D}$ and a learnable pair embedding $\mathbf{e}_\text{pair}(i)\in\mathbb{R}^{D}$ shared by both points in the same correspondence. The final token is
\begin{equation}
    \mathbf{t}_i^v =
    \mathbf{e}_\text{pos}(\bar{\mathbf{p}}_i^v)
    + \mathbf{e}_\text{img}(v)
    + \mathbf{e}_\text{pair}(i)
    \in \mathbb{R}^{D}.
    \label{eq:prompt_token}
\end{equation}
For each view, visible correspondence tokens form $\mathbf{T}_v\in\mathbb{R}^{K_v\times D}$. The tokenizer is shared across both view adapters.

\subsection{Point Correspondence Adapter}
\label{sec:adapter}

We inject correspondence information into DUSt3R's decoder using a two-way attention adapter inspired by SAM~\cite{sam}, enabling bidirectional exchange between sparse correspondence tokens and dense image tokens (\cref{fig:architecture}). 

\paragraph{Two-Way Attention.}
The adapter has $L=2$ layers. Each layer applies: (1) self-attention over prompt tokens $\mathbf{T}_v$; (2) prompt-to-image cross-attention, $\text{Attn}(\mathbf{T}_v,\mathbf{G}_v,\mathbf{G}_v)$; (3) an MLP on prompt tokens; and (4) image-to-prompt cross-attention, $\text{Attn}(\mathbf{G}_v,\mathbf{T}_v,\mathbf{T}_v)$. This bidirectional design lets prompts condition image features with correspondence constraints while allowing prompt tokens to absorb local visual context.

\paragraph{Image Positional Encoding.}
For cross-attention, image positional encodings are computed by applying the same Fourier PE to normalized DUSt3R token-grid positions. Thus, prompt and image tokens share a common coordinate space, allowing cross-attention to compare their spatial locations consistently.

\begin{figure}[t]
  \centering
  \includegraphics[width=\textwidth]{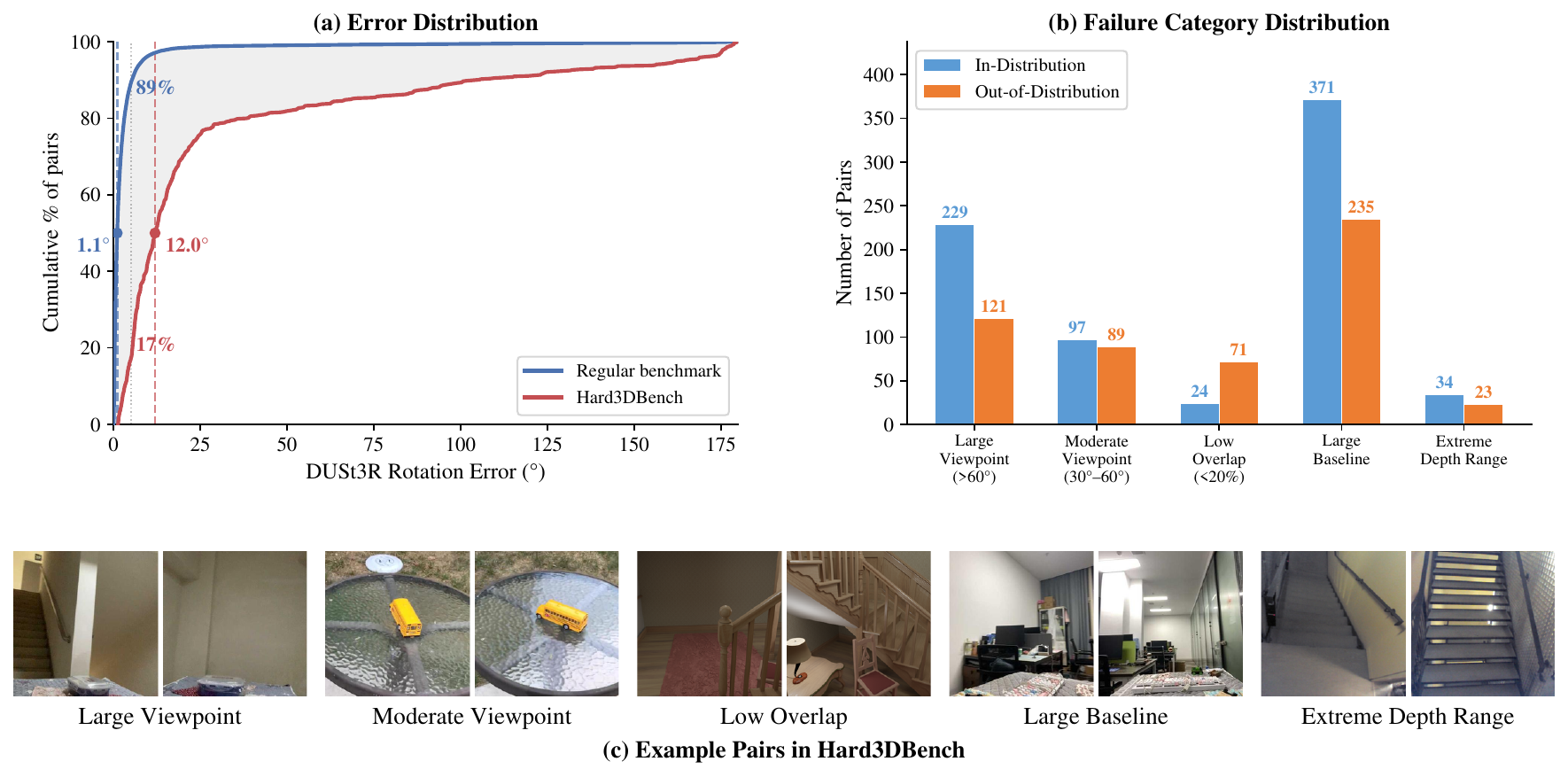}
  \caption{\textbf{Hard3DBench.} Statistics shown for the 2-view stereo subset of Hard3DBench.
  (a)~Hard3DBench produces substantially larger pretrained DUSt3R rotation errors than the regular benchmark.
  (b)~Failure category distribution.
  (c)~Example image pairs in Hard3DBench. 
  }
  \label{fig:hard3dbench}
  \vspace{-3mm}
\end{figure}

\paragraph{Asymmetric Design.}
We employ independent adapters $\mathcal{A}_1, \mathcal{A}_2$ for view~1 and view~2, each with a separate two-way attention adapter but sharing the same correspondence tokenizer:
\begin{align}
    \mathbf{G}_1' &= \mathcal{A}_1(\mathbf{G}_1,\ \mathbf{T}_1,\ \text{pos}_1) \\
    \mathbf{G}_2' &= \mathcal{A}_2(\mathbf{G}_2,\ \mathbf{T}_2,\ \text{pos}_2)
\end{align}
This reflects DUSt3R's asymmetric decoder, where each view has a separate branch.
When no prompts are provided ($K=0$, $\mathbf{T}_v= \varnothing$), the adapter reduces to an identity mapping: $\mathbf{G}_v' = \mathbf{G}_v$.

\subsection{Training Details}
\label{sec:training}

We use DUSt3R's confidence-weighted regression loss:
\begin{equation}
    \mathcal{L} = \sum_{v \in \{1,2\}} \sum_{\mathbf{x}} C_v(\mathbf{x}) \cdot \|\hat{X}_v(\mathbf{x}) - X_v^*(\mathbf{x})\|_2 - \alpha \log C_v(\mathbf{x})
    \label{eq:loss}
\end{equation}
where $C_v$ is predicted confidence, $\hat{X}_v$ and $X_v^*$ are predicted and ground-truth pointmaps, and $\alpha\!=\!0.2$.
The loss is computed in a scale invariant manner.

We generate training prompts from ground-truth 3D geometry by sampling a curriculum over the number of correspondences and keeping only mutually visible correspondences. See more training and prompt generation details in \ref{training_details}.

\section{Hard3DBench}
\label{sec:benchmark}

The primary value of \mmethod{} is on challenging scenarios where DUSt3R fails.
Standard evaluations report only aggregate metrics, masking catastrophic failures on geometrically ambiguous scenes.
To enable systematic analysis, we introduce \textbf{Hard3DBench}, a curated benchmark for both \emph{2-view stereo} and \emph{5-view multi-view} settings, accompanied by a \emph{failure taxonomy} that exposes the specific failure modes of feed-forward 3D reconstruction.
The 2-view setting contains 404 in-distribution and 300 out-of-distribution pairs, and the 5-view setting 400 in-distribution and 300 out-of-distribution tuples.
In-distribution samples come from held-out test/validation splits of four datasets used to train DUSt3R and \mmethod{}: BlendedMVS~\cite{blendedmvs}, CO3D~\cite{co3d}, Habitat~\cite{habitat}, and WildRGB-D~\cite{wildrgbd}.
Out-of-distribution samples come from five datasets absent from training: 7Scenes~\cite{7scenes}, DTU~\cite{dtu}, ETH3D~\cite{eth3d}, NRGBD~\cite{nrgbd}, and ScanNet~\cite{scannet}.
Full curation details are in Appendix~\ref{sec:supp_benchmark}.

\paragraph{Failure Taxonomy.}
We identify five primary failure modes in the 2-view subset (\cref{fig:hard3dbench}): \emph{large viewpoint change} ($>60^\circ$), \emph{moderate viewpoint change} ($30^\circ$--$60^\circ$), \emph{low overlap} ($<20\%$), \emph{large baseline}, and \emph{extreme depth range}. These categories capture geometric conditions that make correspondence estimation and pose recovery difficult, such as reduced visual matches, amplified triangulation error, and increased ambiguity. Since each category is computed from ground-truth poses and depth, challenging pairs can be classified automatically without manual annotation.

\section{Experiments}
\label{sec:experiments}

 \begin{table}[t]                                                                                                                                                            
    \centering                                 
    \caption{\textbf{Standard benchmark results} on held-out splits of CO3D, ScanNet++, and ARKitScenes.
      All methods use $224\!\times\!224$ inputs; {\color{gray}gray rows} are VGGT and Pi3X at both 518 (native) and 224 resolutions, shown only as reference and not included in the bold/underline ranking.
      DUSt3R (retrained) is our re-trained DUSt3R baseline using the same training data as \mmethod{} for fair comparison;
      {\tiny\color{teal}$\pm$x.x} denotes the mAA@30 difference from this baseline.
      PM Err.: median pointmap error.
      Best in \textbf{bold}, second best \underline{underlined}.}
    \label{tab:main}
    \small
    \setlength{\tabcolsep}{4pt}
    \begin{tabular}{@{}l cc cc cc@{}}
      \toprule
      & \multicolumn{2}{c}{CO3D} & \multicolumn{2}{c}{ScanNet++} & \multicolumn{2}{c}{ARKitScenes} \\
      \cmidrule(lr){2-3} \cmidrule(lr){4-5} \cmidrule(lr){6-7}
      Method & mAA@30\textdegree\,$\uparrow$ & PM Err.\,$\downarrow$ & mAA@30\textdegree\,$\uparrow$ & PM Err.\,$\downarrow$ & mAA@30\textdegree\,$\uparrow$ & PM Err.\,$\downarrow$ \\
      \midrule
      {\color{gray} VGGT~\cite{vggt} (518)} & {\color{gray} 87.76} & {\color{gray} 0.498} & {\color{gray} 59.47} & {\color{gray} 0.261} & {\color{gray} 77.91} & {\color{gray} 0.159} \\
      {\color{gray} Pi3X~\cite{pi3} (518)} & {\color{gray} 87.95} & {\color{gray} 0.467} & {\color{gray} 80.77} & {\color{gray} 0.140} & {\color{gray} 82.16} & {\color{gray} 0.146} \\
      {\color{gray} VGGT~\cite{vggt} (224)} & {\color{gray} 62.53} & {\color{gray} 1.994} & {\color{gray} 24.47} & {\color{gray} 1.325} & {\color{gray} 40.89} & {\color{gray} 0.690} \\
      {\color{gray} Pi3X~\cite{pi3} (224)} & {\color{gray} 85.02} & {\color{gray} 0.577} & {\color{gray} 72.53} & {\color{gray} 0.196} & {\color{gray} 76.39} & {\color{gray} 0.167} \\[2pt]
      DUSt3R~\cite{dust3r}               & 79.20 & 0.728 & 82.57 & 0.190 & \underline{86.02} & \underline{0.146} \\
      MASt3R~\cite{mast3r}               & 78.45 & 0.680 & 51.79 & 0.459 & 65.40 & 0.272 \\
      \cdashline{1-7}\\[-8pt]
      DUSt3R (retrained)                 & 84.33 & 0.659 & 90.81 & 0.174 & 82.71 & 0.155 \\
      \mmethod{} w/o prompts             & 84.41{\tiny\color{teal}+0.1} & 0.678 & 89.87{\tiny\color{teal}$-$0.9} & 0.175 & 84.27{\tiny\color{teal}+1.6} & 0.157 \\
      \mmethod{} (1p\_best5)             & \underline{85.71}{\tiny\color{teal}+1.4} & \textbf{0.651} & \underline{91.75}{\tiny\color{teal}+0.9} & \textbf{0.167} & 85.90{\tiny\color{teal}+3.2} & \textbf{0.141} \\
      \mmethod{} (grid oracle)           & \textbf{88.21}{\tiny\color{teal}+3.9} & \underline{0.658} & \textbf{93.94}{\tiny\color{teal}+3.1} & \underline{0.169} & \textbf{92.75}{\tiny\color{teal}+10.0} & \underline{0.146} \\
      \bottomrule
    \end{tabular}
    \vspace{-5mm}
  \end{table}

\subsection{Evaluation Setup}
\label{sec:exp_setup}

\paragraph{Baselines.}
We compare against state-of-the-art pairwise 3D reconstruction methods: (1)~DUSt3R~\cite{dust3r}, the original pretrained model; (2)~DUSt3R (re-trained), the same architecture re-trained on the same data as \mmethod{} to isolate the effect of prompting from additional training data; (3)~MASt3R~\cite{mast3r}, which extends DUSt3R with dense local feature matching heads; (4)~VGGT~\cite{vggt}, a feed-forward vision foundation model for 3D geometry tasks; and (5)~Pi3/Pi3X~\cite{pi3}, a permutation-equivariant multi-view reconstruction model.
VGGT and Pi3/Pi3X use larger architectures and are trained on much more data, including proprietary internal datasets, with much higher computational budgets. We therefore include them as reference rather than apples-to-apples baselines.

\paragraph{Datasets.}
For standard evaluation, we use held-out splits of three benchmarks: CO3D~\cite{co3d}, ScanNet++~\cite{scannetpp}, and ARKitScenes~\cite{arkitscenes}. For each dataset, we sample 1{,}000 image pairs with a fixed random seed and use the same pairs across all methods. We additionally evaluate on the Hard3DBench benchmark (\cref{sec:hard_cases}).

\paragraph{Metrics.}
Following DUSt3R~\cite{dust3r}, we report two metrics: scale-and-shift-invariant pointmap error for 3D reconstruction quality (lower is better) and mean Average Accuracy at threshold $\theta^\circ$ (mAA@$\theta$, higher is better) for relative camera pose accuracy. Tables report mAA@30\textdegree; results at 5\textdegree{} and 15\textdegree{} are in Appendix~\ref{sec:supp_strict}. Implementation details are in Appendix~\ref{sec:supp_details}.

\paragraph{Prompt selection strategies.}
We evaluate several prompt selection strategies. (a) \textbf{Random oracle} ($K$p\_best$B$): sample $K$ correspondences $B$ independent times and select the set that minimizes reconstruction loss. If there's no best$B$ mentioned, $K$ correspondences are sampled randomly. (b) \textbf{Grid oracle}: partition view~1 into a $16\times16$ grid of 14-pixel cells ($\sim$256 cells), select one random correspondence per populated cell, run a batched forward pass for each candidate, and keep the single correspondence whose cell yields the greatest pose improvement (reduction in $\max(R_\mathrm{err},T_\mathrm{err})$) over the no-prompt baseline. Samples without valid correspondences fall back to the no-prompt result.

\subsection{Evaluation on Standard Benchmark}
\label{sec:main_results}

\cref{tab:main} presents results on standard benchmarks. Without prompts, \mmethod{} performs on par with DUSt3R (retrained). Adding even a single well-chosen prompt (1p\_best5) consistently improves over the unprompted baseline. The grid oracle reveals substantially larger headroom, with a striking $+8.5\%$ mAA@30° on ARKitScenes (from 84.27\% to 92.75\%). This confirms that \emph{prompt location matters significantly}: the model can leverage even a single correspondence to meaningfully reduce pose error when placed in the right region. Notably, the grid oracle delivers the largest pose gains while pointmap error improves only modestly, indicating that prompts primarily reshape the overall scene configuration rather than uniformly tightening local surface fits.

\begin{table}[t]
    \centering
    \caption{\textbf{Hard3DBench results} on the 2-view (left) and 5-view (right) subsets.
      In-distribution samples come from held-out splits of datasets used to train DUSt3R and \mmethod{}; out-of-distribution samples come from datasets absent from training.
      \mmethod{} K denotes $K$ GT correspondences as prompts (per pairwise forward in the 5-view setting).
      For 5-view evaluation, all DUSt3R-family methods (including \mmethod{}) follow DUSt3R's global alignment protocol over the C(5,2) pairwise predictions.
      Comparison methods use $224\!\times\!224$ resolution except MASt3R (512); {\color{gray}gray rows} are VGGT and Pi3 at both 518 (native) and 224 resolutions, shown only as reference and not included in the bold/underline ranking.
      Methods below the dashed line share identical training data;
      {\tiny\color{teal}+x.x} denotes mAA improvement over DUSt3R (retrained).
      Pointmap Error is the median across pairs/tuples.
      Best in \textbf{bold}, second best \underline{underlined}.}
    \label{tab:hard}
    \small
    \setlength{\tabcolsep}{4pt}
    \begin{tabular}{@{}l cc cc@{}}
      \toprule
      & \multicolumn{2}{c}{\textit{2-view (pair)}} & \multicolumn{2}{c}{\textit{5-view (tuple)}} \\
      \cmidrule(lr){2-3} \cmidrule(lr){4-5}
      Method
        & mAA@30\textdegree\,$\uparrow$ & Pointmap Err.\,$\downarrow$
        & mAA@30\textdegree\,$\uparrow$ & Pointmap Err.\,$\downarrow$ \\
      \midrule
      \multicolumn{5}{@{}l}{\textit{In-Distribution (404 pairs / 400 tuples)}} \\[2pt]
      {\color{gray} VGGT~\cite{vggt} (518)}    & {\color{gray} 63.7} & {\color{gray} 0.256} & {\color{gray} 56.5} & {\color{gray} 1.222} \\
      {\color{gray} Pi3~\cite{pi3} (518)}      & {\color{gray} 64.5} & {\color{gray} 0.232} & {\color{gray} 60.3} & {\color{gray} 0.829} \\
      {\color{gray} VGGT~\cite{vggt} (224)}    & {\color{gray} 40.5} & {\color{gray} 0.620} & {\color{gray} 33.2} & {\color{gray} 1.536} \\
      {\color{gray} Pi3~\cite{pi3} (224)}      & {\color{gray} 60.7} & {\color{gray} 0.299} & {\color{gray} 56.9} & {\color{gray} 1.234} \\[2pt]
      DUSt3R~\cite{dust3r}                     & 39.2 & 0.502 & 21.1 & 2.251 \\
      MASt3R~\cite{mast3r} (512)               & \textbf{62.9} & 0.229 & 48.4 & \textbf{1.333} \\
      \cdashline{1-5}\\[-8pt]
      DUSt3R (retrained)                       & 56.1 & 0.216 & 39.4 & 2.131 \\
      \mmethod{} w/o prompts                   & 55.3 & 0.244 & 38.8 & 2.121 \\
      \mmethod{} (1 prompt)                    & 57.5{\tiny\color{teal}+1.3} & 0.218 & 45.0{\tiny\color{teal}+5.6} & 1.883 \\
      \mmethod{} (4 prompts)                   & 61.3{\tiny\color{teal}+5.2} & \underline{0.196} & \underline{48.6}{\tiny\color{teal}+9.2} & 1.584 \\
      \mmethod{} (8 prompts)                   & \underline{61.8}{\tiny\color{teal}+5.7} & \textbf{0.186} & \textbf{49.3}{\tiny\color{teal}+9.9} & \underline{1.525} \\
      \midrule
      \multicolumn{5}{@{}l}{\textit{Out-of-Distribution (300 pairs / 300 tuples)}} \\[2pt]
      {\color{gray} VGGT~\cite{vggt} (518)}    & {\color{gray} 81.3} & {\color{gray} 0.329} & {\color{gray} 81.1} & {\color{gray} 0.099} \\
      {\color{gray} Pi3~\cite{pi3} (518)}      & {\color{gray} 80.6} & {\color{gray} 0.318} & {\color{gray} 81.4} & {\color{gray} 0.096} \\
      {\color{gray} VGGT~\cite{vggt} (224)}    & {\color{gray} 47.4} & {\color{gray} 1.766} & {\color{gray} 47.1} & {\color{gray} 0.534} \\
      {\color{gray} Pi3~\cite{pi3} (224)}      & {\color{gray} 71.1} & {\color{gray} 0.433} & {\color{gray} 73.5} & {\color{gray} 0.143} \\[2pt]
      DUSt3R~\cite{dust3r}                     & 45.3 & 0.751 & 54.7 & \underline{0.253} \\
      MASt3R~\cite{mast3r} (512)               & 51.6 & \textbf{0.528} & 58.0 & \textbf{0.225} \\
      \cdashline{1-5}\\[-8pt]
      DUSt3R (retrained)                       & 52.6 & 0.700 & 54.8 & 0.274 \\
      \mmethod{} w/o prompts                   & 50.8 & 0.781 & 54.7 & 0.310 \\
      \mmethod{} (1 prompt)                    & 56.6{\tiny\color{teal}+4.0} & 0.689 & 56.7{\tiny\color{teal}+1.9} & 0.268 \\
      \mmethod{} (4 prompts)                   & \underline{60.4}{\tiny\color{teal}+7.8} & \underline{0.683} & \underline{58.6}{\tiny\color{teal}+3.8} & 0.263 \\
      \mmethod{} (8 prompts)                   & \textbf{62.4}{\tiny\color{teal}+9.8} & 0.686 & \textbf{60.0}{\tiny\color{teal}+5.2} & 0.254 \\
      \bottomrule
    \end{tabular}
    \vspace{-5mm}
  \end{table}

\subsection{Evaluation on Hard3DBench}
\label{sec:hard_cases}

\begin{figure}[t]
  \centering
  \includegraphics[width=\textwidth]{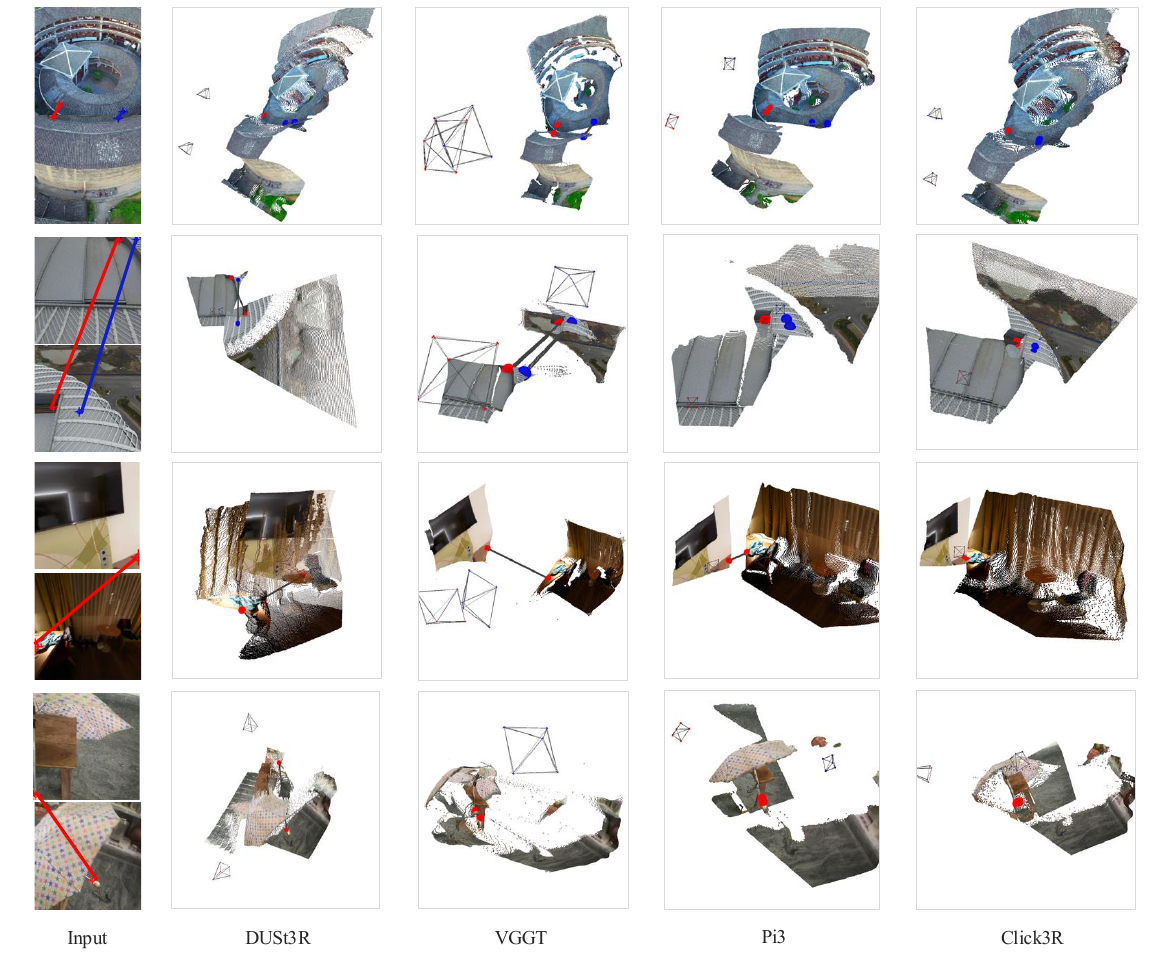}
  \caption{\textbf{Qualitative comparison.}
Each example features challenging geometry, such as large baselines, limited co-visibility, or substantial viewpoint changes. The input image pair with user-provided correspondence prompts (\textit{colored dots}) is shown on the left, followed by each model's reconstruction. \mmethod{} produces more accurate and consistent 3D reconstructions, with prompted correspondences aligning more closely in 3D. Additional qualitative results are provided in Appendix~\cref{fig:supp_qualitative}.}
  \label{fig:qualitative}
  \vspace{-6mm}
\end{figure}

\cref{tab:hard} reports results on both the 2-view and 5-view subsets.
\mmethod{} consistently improves mAA@30° over the retrained DUSt3R baseline across all four conditions, with the largest gain of $+9.9$ on the 5-view in-distribution split, and pointmap error also improving in all four conditions.
Notably, these improvements are obtained without increasing input resolution or training data.
These results suggest that even sparse correspondence prompts effectively resolve failure modes in highly ambiguous scenes, enabling the model to recover accurate geometry where fully automatic methods struggle.
\cref{fig:qualitative} shows qualitative comparisons: a few correspondence clicks placed in geometrically ambiguous regions resolve failures that produce dramatically incorrect geometry in DUSt3R.
See Per-Category Analysis in Appendix~\cref{par:per-category_analysis}.

\subsection{Evaluation with Human-Labeled Correspondences}
\label{sec:human_eval}

The previous experiments use ground-truth correspondences derived from known 3D geometry. To evaluate \mmethod{} under realistic annotation conditions, we test on 100 image pairs with human-labeled correspondences. The pairs are drawn from out-of-distribution set of Hard3DBench and cover challenging geometric configurations. A human annotator, not specifically familiar with the evaluated datasets or scenes, freely chooses sparse correspondences (3.8 per pair on average, from 1 to 7) via a labeling interface (Figure~\ref{fig:labeling_tool}), without access to ground-truth depth or camera parameters.

\cref{tab:human} demonstrates that human-provided prompts drastically reduce median rotation error compared to the unprompted model (8.78\textdegree{} vs.\ 17.30\textdegree{}) and raise mAA@30 from 36.2 to 56.2, along with significant improvement in Relative Rotation Accuracy. Compared to the random ground-truth oracle ($K\!=\!4$), human labels recover about 96\% of the medRRE improvement gap, showing that sparse, imperfect human clicks capture most of the benefit of perfect correspondences. These results confirm that \mmethod{} translates effectively from synthetic oracle prompts to practical human interaction.

\begin{table}[t]
    \centering
    \caption{\textbf{Human-labeled correspondence evaluation.}
      100 out-of-distribution pairs with human-annotated correspondences.
      \mmethod{} (human) uses sparse clicks from a human annotator;
      \mmethod{} (random GT) uses $K\!=\!4$ randomly sampled ground-truth correspondences as an oracle upper bound.
      {\tiny\color{teal}+x.x} denotes improvement over DUSt3R (retrained).
      Best in \textbf{bold}, second best \underline{underlined}.}
    \label{tab:human}
    \small
    \setlength{\tabcolsep}{3pt}
    \begin{tabular}{@{}l cccccc@{}}
      \toprule
      Method
        & medRRE\,$\downarrow$
        & RRA@5\textdegree\,$\uparrow$
        & RRA@15\textdegree\,$\uparrow$
        & mAA@5\,$\uparrow$
        & mAA@15\,$\uparrow$
        & mAA@30\,$\uparrow$ \\
      \midrule
      DUSt3R~\cite{dust3r}       & 24.67\textdegree & 1.0\% & 37.0\% & 0.0 & 9.5 & 25.6 \\
      \cdashline{1-7}\\[-8pt]
      DUSt3R (retrained)         & 15.05\textdegree & 10.0\% & 50.0\% & 2.4 & 18.8 & 35.7 \\
      \mmethod{} w/o prompts     & 17.30\textdegree & 13.0\% & 47.0\% & 4.6 & 21.5 & 36.2 \\
      \mmethod{} (human)         & \underline{8.78\textdegree} & \underline{22.0\%}{\tiny\color{teal}+12.0} & \underline{69.0\%}{\tiny\color{teal}+19.0} & \underline{9.2}{\tiny\color{teal}+6.8} & \underline{37.3}{\tiny\color{teal}+18.5} & \underline{56.2}{\tiny\color{teal}+20.5} \\
      \mmethod{} (random GT $K\!=\!4$) & \textbf{8.40\textdegree} & \textbf{27.0\%}{\tiny\color{teal}+17.0} & \textbf{79.0\%}{\tiny\color{teal}+29.0} & \textbf{9.4}{\tiny\color{teal}+7.0} & \textbf{39.5}{\tiny\color{teal}+20.7} & \textbf{59.2}{\tiny\color{teal}+23.5} \\
\bottomrule
    \end{tabular}
  \end{table}

\subsection{Ablation Study}

\paragraph{Training strategy ablation.}
We compare three training strategies for \mmethod{}: (i) freezing the DUSt3R backbone and training only the adapter, (ii) finetuning DUSt3R jointly with the adapter, and (iii) training the full pipeline from CroCo v2~\cite{crocov2} initialization. Averaged over the three standard benchmarks and both Hard3DBench 2-view splits with one GT click (Appendix~\ref{sec:supp_trainstrat}), training from CroCo v2 performs best (mAA@30 75.1\%), followed by DUSt3R finetuning (72.3\%), while freezing DUSt3R gives the weakest results (71.2\%). This suggests that \mmethod{} requires substantial end-to-end adaptation to align backbone features with correspondence-conditioned reasoning; training the adapter alone is insufficient.

\begin{figure}[t]
  \centering
  \begin{subfigure}{0.49\linewidth}
    \centering
    \includegraphics[width=\linewidth]{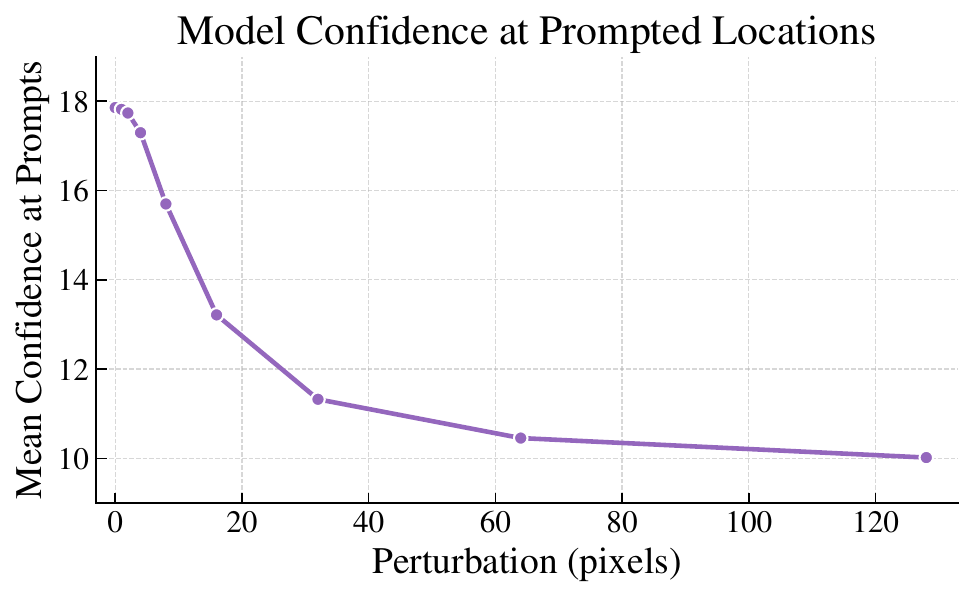}
    \caption{Confidence drops as perturbation magnitude increases.}
    \label{fig:perturb_confidence}
  \end{subfigure}
  \hfill
  \begin{subfigure}{0.49\linewidth}
    \centering
    \includegraphics[width=\linewidth]{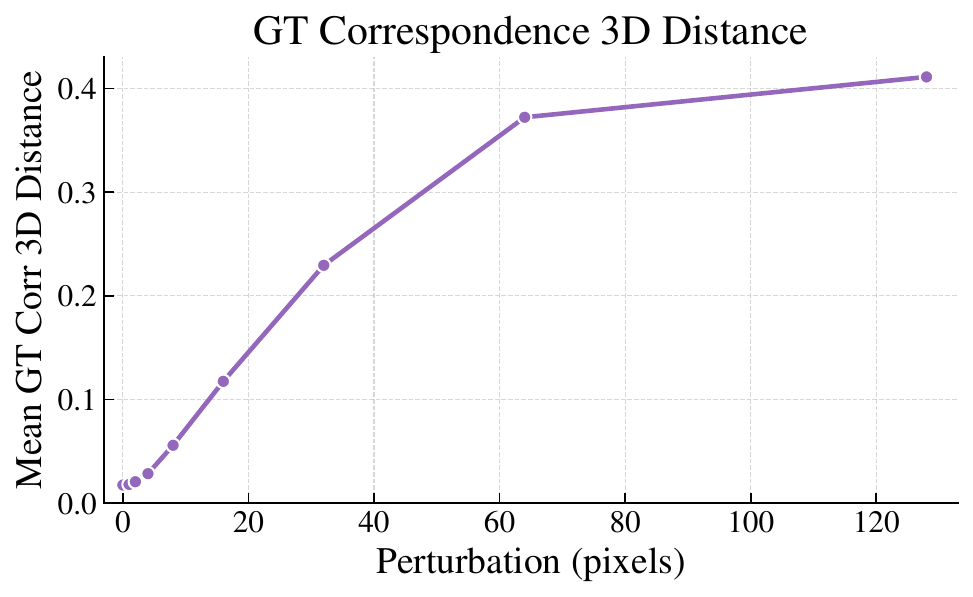}
    \caption{Misleading hints impacting model matching performance}
    \label{fig:perturb_dataset_arkit}
  \end{subfigure}

  \caption{\textbf{Correspondence perturbation analysis.}
  (a) Model confidence at the clicked pixels decreases as perturbation strength increases (CO3D, ScanNet++, and ARKitScenes).
  (b) 3D distance between the predicted points at the ground-truth correspondence (ARKitScenes): the model remains robust under small perturbations, but larger perturbations degrade matching performance.}
  \label{fig:corr_perturb}
  \vspace{-5mm}
\end{figure}

\paragraph{Correspondence perturbation analysis.}
To study how the model uses sparse correspondence hints, we perturb a single correspondence by fixing the click in view~1 and shifting its matched point in view~2 along a random direction. As perturbation increases, model confidence consistently decreases, indicating sensitivity to correspondence reliability (Fig.~\ref{fig:perturb_confidence}).

We further measure whether predictions follow the true or perturbed correspondence by computing the 3D distance between predicted points at the ground-truth correspondence locations in both views. The model remains largely robust to small perturbations, but becomes more sensitive once perturbations reach half a patch (8 pixels), leading to larger degradation (Fig.~\ref{fig:perturb_dataset_arkit}). For more analysis on how correspondences affects the adapters and downstream tasks, see \cref{sec:supp_adapter}.

We also refer readers to our analyses of the effect of the number of prompts (\cref{sec:supp_k_ablation}), robustness to image augmentation (\cref{sec:supp_augmentation}), and error distribution (\cref{sec:supp_distribution}).

\section{Conclusion}
\label{sec:conclusion}

We presented \mmethod{}, an interactive stereo 3D reconstruction framework that bridges fully automatic feed-forward methods, correspondence-based classical pipelines, and recent prompt-based vision systems.
By injecting sparse point correspondences into a feed-forward stereo 3D reconstruction model through a lightweight correspondence adapter, \mmethod{} enables users to guide the reconstruction process without test-time model fine-tuning or optimization, resolving geometric ambiguities that often cause catastrophic failures.
Experiments on our curated Hard3DBench demonstrate that even a handful of correspondences can dramatically improve reconstruction quality in challenging scenarios.
While \mmethod{} has limitations (Appendix~\ref{sec:limitations}) and failure cases (Appendix~\ref{sec:supp_failures}), we hope this work inspires further exploration of correspondence prompting as a promising direction for interactive 3D reconstruction, analogous to how SAM~\cite{sam} advanced interactive prompting for 2D segmentation.

\begin{ack}
The authors acknowledge the University of Virginia Research Computing and Data Analytics Center, Advanced Micro Devices AI and HPC Cluster Program, Advanced Cyberinfrastructure Coordination Ecosystem: Services \& Support (ACCESS) program, and National Artificial Intelligence Research Resource (NAIRR) Pilot for computational resources, including the Anvil supercomputer (National Science Foundation award OAC 2005632) at Purdue University and the Delta and DeltaAI advanced computing resources (National Science Foundation award OAC 2005572).
The authors would also like to thank Subhransu Maji for insightful discussions.
\end{ack}

{
\small
\bibliographystyle{plainnat}
\bibliography{main}
}

\newpage
\clearpage
\appendix

\renewcommand{\thesection}{\Alph{section}}
\setcounter{section}{0}

\section*{Supplementary Material}

We provide implementation details and training configuration (\cref{sec:supp_details}), details on Hard3DBench construction (\cref{sec:supp_benchmark}), per-category analysis on the 2-view subset (\cref{par:per-category_analysis}), additional experiments including training strategy ablation on all evaluation sets, stricter pose thresholds, prompt count ablation, augmentation robustness, incorrect prompts, automatic prompts, MASt3R transfer, and error distribution analysis (\cref{sec:supp_experiments}), an internal analysis of how the correspondence adapter modifies encoder features and decoder attention (\cref{sec:supp_adapter}), additional qualitative results (\cref{sec:supp_qualitative}), failure case analysis (\cref{sec:supp_failures}), licenses of the existing assets we use (\cref{sec:supp_licenses}), limitation (\cref{sec:limitations}), and broader impacts (\cref{sec:supp_broader_impacts}).

\section{Implementation Details}
\label{sec:supp_details}

\mmethod{} follows the DUSt3R~\cite{dust3r} training recipe (loss function, optimizer, data augmentation, symmetric batching) and inherits its CroCo v2~\cite{crocov2} ViT-Large encoder and dual asymmetric decoder.
Below we detail the components specific to \mmethod{}.

\paragraph{Correspondence adapter.}
We insert two SAM-style~\cite{sam} Two-Way Transformer adapters (depth 2, 8 heads, MLP dim 2048, embedding dim 768) between the encoder and decoder (\cref{sec:adapter}).
The two adapters share a single correspondence tokenizer (\cref{sec:prompt_encoder}) that encodes each correspondence point as the sum of a positional encoding (random Fourier features~\cite{sam}), an image identity embedding, and a pair identity embedding (vocabulary size 256).
When prompts are provided, the adapted features replace the encoder output; otherwise, the adapter is bypassed and the model reduces to standard DUSt3R.
The adapters and tokenizer are randomly initialized; all other weights are initialized from the CroCo pretrained checkpoint.
All parameters are trained jointly end-to-end.

\paragraph{Training data.}
We train on nine datasets, each contributing 100{,}000 image pairs:
BlendedMVS~\cite{blendedmvs},
MegaDepth~\cite{megadepth},
CO3D~\cite{co3d},
StaticThings3D~\cite{staticthings3d},
Waymo Open~\cite{xu2025wod},
ARKitScenes~\cite{arkitscenes},
WildRGB-D~\cite{wildrgbd},
Habitat~\cite{habitat}, and
ScanNet++~\cite{scannetpp}.
Validation uses held-out splits from six of these datasets with 1{,}000 pairs each.

\paragraph{Ground-truth prompt generation.}
During training, correspondence prompts are generated by projecting view~1's 3D pointmap into view~2 and accepting correspondences where the 3D distance to the nearest view~2 surface point is below 5\,cm (which implicitly rejects occluded points).
Per sample, 30\% receive zero prompts (adapter bypassed), 40\% receive 1--4, and 30\% receive 5--12 correspondences.

\paragraph{Training configuration.}
We train for 100 epochs on 4 GPUs with per-GPU batch size 16, using AdamW ($\beta_1{=}0.9$, $\beta_2{=}0.95$, weight decay $0.05$), a learning rate of $10^{-4}$ with 10 epochs of linear warmup followed by cosine decay to $10^{-6}$, and float16 mixed precision.

\paragraph{Baseline image preprocessing.}
DUSt3R-family models (DUSt3R, MASt3R, \mmethod{}) use images normalized to $[-1,1]$. The VGGT and Pi3 wrappers internally denormalize images to $[0,1]$ before forwarding to match each model's expected input range.

\paragraph{Standard benchmark data.}
The held-out splits are the CO3D test sequences, the ScanNet++ validation scenes, and the ARKitScenes validation videos, each processed with the DUSt3R pipeline.

\paragraph{Evaluation metric details.}
The \textbf{scale-and-shift-invariant pointmap error} uses the normalization of DUSt3R: we divide the predicted pointmap by its mean point distance, shift both pointmaps by their median depth, and rescale the prediction to the median distance of the ground-truth points to their median point. We then average the per-pixel L2 distance over all valid pixels (i.e., L2 within each 3D point, mean across pixels).
For \textbf{mAA@$\theta$}, relative camera pose is first estimated from each model's predicted pointmaps via PnP-RANSAC (\texttt{cv2.solvePnPRansac} with the SQPNP solver), recovering per-view camera-to-world poses.
Rotation error is the geodesic distance $\arccos\!\big((\mathrm{tr}(R_\mathrm{gt}^\top R_\mathrm{pred})-1)/2\big)$ in degrees; translation error is the angular deviation between predicted and ground-truth translation directions, with sign ambiguity handled by $\min(\alpha,180^\circ-\alpha)$.
mAA@$\theta$ is the area under the accuracy curve of the pose error $\max(R_\mathrm{err},T_\mathrm{err})$ up to threshold $\theta^\circ$.
All pairs are \emph{symmetrized} (evaluated in both directions) following the standard DUSt3R protocol.

\paragraph{Runtime and Model Complexity.}
\cref{tab:runtime} reports parameter counts and GPU compute times on a single NVIDIA A100 at $224\!\times\!224$ resolution using bf16 mixed precision.
The correspondence adapter adds 34.1M parameters (6.4\%) over DUSt3R.
When no prompts are provided, the adapter is bypassed entirely and \mmethod{} has the same compute cost as DUSt3R (20.4\,ms vs.\ 20.4\,ms).
With $K\!=\!8$ prompts, the two-way transformer adds ${\sim}1.0$\,ms of GPU compute (5\%), and this overhead is nearly constant across prompt counts since the bottleneck is the fixed-size image token processing, not the number of prompts.
All times report pure GPU compute (CUDA kernel time) for the model forward pass only, excluding prompt acquisition.

\begin{table}[t]
  \centering
  \caption{\textbf{Runtime and model complexity.}
    GPU compute time measured on a single A100 ($224\!\times\!224$, bf16).
    \mmethod{} without prompts has identical cost to DUSt3R; $K\!=\!8$ prompts add 5\% overhead.}
  \label{tab:runtime}
  \small
  \setlength{\tabcolsep}{4pt}
  \begin{tabular}{@{}lrrr@{}}
    \toprule
    Configuration & Params (M) & GPU time (ms) & Overhead \\
    \midrule
    DUSt3R~\cite{dust3r} (pretrained) & 532.3 & 20.4 & --- \\
    \mmethod{} (no prompts)           & 566.4 & 20.4 & $+$0\% \\
    \mmethod{} ($K\!=\!8$)            & 566.4 & 21.5 & $+$5\% \\
    \mmethod{} ($K\!=\!12$)           & 566.4 & 21.4 & $+$5\% \\
    \bottomrule
  \end{tabular}
\end{table}

\section{Training Details}
\label{training_details}
\paragraph{Generating Training Correspondences.}
During training, we generate prompts from ground-truth 3D geometry.
For each image pair with known cameras and pointmaps, we first sample the number of correspondence prompts according to a curriculum: 30\% of samples use no prompts ($K=0$), 40\% use a small number of prompts ($K \in [1,4]$), and 30\% use more prompts ($K \in [5,12]$). Given the sampled value of $K$, we:
(1)~sample $K$ 3D points from the first view;
(2)~project them into both views using known intrinsics and extrinsics;
(3)~retain only points that are mutually visible, verified by comparing the projected 3D position against the second view's GT pointmap within a distance threshold.

\paragraph{Training Protocol.}
We initialize from CroCo v2~\cite{crocov2} pretrained weights and train all parameters end-to-end, including both the base model and the adapter.
Training uses 4 GPUs, AdamW ($\text{lr}\!=\!10^{-4}$) with linear warmup and cosine decay.
We train on the same dataset mixture as DUSt3R: CO3D~\cite{co3d}, ARKitScenes~\cite{arkitscenes}, ScanNet++~\cite{scannetpp}, WildRGB-D~\cite{wildrgbd}, BlendedMVS~\cite{blendedmvs}, MegaDepth~\cite{megadepth}, Habitat~\cite{habitat}, Waymo~\cite{xu2025wod}, and StaticThings3D~\cite{staticthings3d} (100K pairs each), at $224 \times 224$ resolution.
With batch size 16 per GPU on A100s, a training step takes 0.60\,s for \mmethod{} and 0.55\,s for DUSt3R (retrained), which amounts to about 234 and 215 hours for 100 epochs on 4 GPUs, with a peak memory of 29{,}069 and 28{,}731 MiB per GPU.

\section{Hard3DBench Details}
\label{sec:supp_benchmark}

Hard3DBench has two subsets: a \emph{2-view stereo} subset of 404 in-distribution and 300 out-of-distribution pairs, and a \emph{5-view multi-view} subset of 400 in-distribution and 300 out-of-distribution tuples.
Both subsets use \emph{pretrained} DUSt3R's pose error as the difficulty signal, rather than \mmethod{} or our retrained baseline, so the benchmark is defined independently of the methods under evaluation.
They share the same source datasets: the in-distribution split uses held-out test/val splits of four DUSt3R training datasets, while the out-of-distribution split uses five datasets entirely absent from training (\cref{tab:supp_datasets} reports the 2-view per-dataset statistics).
DTU uses the cameras of MVSNet~\cite{yao2018mvsnet} and the depth maps rendered by Spann3R~\cite{spann3r}, 7Scenes depth is registered to the color camera as in SimpleRecon~\cite{sayed2022simplerecon}, ScanNet is prepared following MonST3R~\cite{monst3r}, and ETH3D following Pi3~\cite{pi3}.

\paragraph{2-view subset.}
To obtain hard cases, we evaluate the pretrained DUSt3R model on up to 20{,}000 randomly drawn image pairs per dataset (${\sim}$69K in-distribution, ${\sim}$68K out-of-distribution), pre-filtering pairs with insufficient co-visibility via depth reprojection in DUSt3R's $224\!\times\!224$ crop space.
We rank pairs by rotation error and select the top-$N$ worst per dataset ($N\!=\!1{,}000$ for BlendedMVS, CO3D, WildRGB-D, DTU, and ETH3D; $N\!=\!300$ for 7Scenes, NRGBD, and ScanNet). For Habitat, only the 406 of its top 1{,}000 pairs from scenes not used in training are kept. This yields 3{,}406 in-distribution and 2{,}900 out-of-distribution candidates.
Each candidate is then classified into one or more geometric failure categories (\cref{tab:supp_taxonomy}) from ground-truth poses and depth.
We uniformly subsample 404 in-distribution and 300 out-of-distribution pairs with a fixed seed.

\paragraph{5-view subset.}
For each scene, we sample up to 30 5-frame tuples by combining stride-based windows (strides 1, 2, 5, 10, 20) with random index sampling and deduplication; tuples are kept only if all five views form a connected graph under pairwise depth-reprojection overlap $\geq 5\%$.
We then run DUSt3R on all $C(5,2){=}10$ symmetrized pairs and compose them via the standard global alignment optimizer (300 iterations, msp initialization), yielding ${\sim}16\text{K}$ in-distribution and ${\sim}4\text{K}$ out-of-distribution candidate tuples.
We rank tuples by mean pairwise rotation error and select the top-100 per in-distribution dataset (400 total) and top-60 per out-of-distribution dataset (300 total).

\paragraph{Difficulty for recent models.}
Compared with random pairs from the same datasets with the same per-dataset counts, VGGT and Pi3 at 518 resolution drop in RRA@5 from 82.6\% and 85.1\% to 54.7\% and 57.9\% on the in-distribution pairs of the 2-view subset, and from 98.2\% and 97.0\% to 88.7\% and 81.3\% on the out-of-distribution pairs, confirming that Hard3DBench remains challenging for more recent 3D reconstruction models.

\begin{table}[t]
  \centering
  \caption{\textbf{Failure taxonomy for Hard3DBench.}
    Categories are multi-label, detected from ground-truth poses and depth.}
  \label{tab:supp_taxonomy}
  \small
  \setlength{\tabcolsep}{4pt}
  \begin{tabular}{@{}ll@{}}
    \toprule
    Category & Detection criterion \\
    \midrule
    Large viewpoint      & Relative rotation $> 60^\circ$ \\
    Moderate viewpoint   & $30^\circ <$ relative rotation $\leq 60^\circ$ \\
    Low overlap          & Co-visible pixel fraction $< 20\%$ \\
    Large baseline       & Baseline\,/\,median depth $> 0.3$ \\
    Extreme depth range  & $d_\mathrm{max}/d_\mathrm{min} > 50$ \\
    \bottomrule
  \end{tabular}
\end{table}

\begin{table}[t]
  \centering
  \caption{\textbf{Dataset composition of the 2-view subset of Hard3DBench.}
    $N_\text{pool}$: pairs after overlap filtering;
    $N_\text{hard}$: candidates selected by worst rotation error;
    $N_\text{final}$: pairs after uniform subsampling.
    The 5-view subset uses the same datasets; per-dataset tuple counts are uniform (top-100 in-distribution, top-60 out-of-distribution).}
  \label{tab:supp_datasets}
  \footnotesize
  \setlength{\tabcolsep}{3.5pt}
  \begin{tabular}{@{}llrrrl@{}}
    \toprule
    Dataset & Split & $N_\text{pool}$ & $N_\text{hard}$ & $N_\text{final}$ & Scene type \\
    \midrule
    \multicolumn{6}{@{}l}{\textit{In-Distribution (404 pairs, uniformly sampled from 3{,}406 candidates)}} \\[2pt]
    BlendedMVS~\cite{blendedmvs}  & val  & 19{,}719 & 1{,}000 &  97 & Synthetic multi-view \\
    CO3D~\cite{co3d}              & test & 19{,}396 & 1{,}000 & 109 & Object-centric video \\
    Habitat~\cite{habitat}        & val  &  9{,}871 &     406 & 100 & Synthetic indoor \\
    WildRGB-D~\cite{wildrgbd}     & test & 19{,}860 & 1{,}000 &  98 & RGB-D video \\
    \midrule
    \multicolumn{6}{@{}l}{\textit{Out-of-Distribution (300 pairs, uniformly sampled from 2{,}900 candidates)}} \\[2pt]
    7Scenes~\cite{7scenes}        & all  & 19{,}690 &   300 &  47 & Indoor relocalization \\
    DTU~\cite{dtu}                & all  &  9{,}141 & 1{,}000 & 106 & Multi-view objects \\
    ETH3D~\cite{eth3d}            & all  &  2{,}162 & 1{,}000 &  81 & Indoor/outdoor \\
    NRGBD~\cite{nrgbd}            & all  & 19{,}669 &   300 &  34 & RGB-D scenes \\
    ScanNet~\cite{scannet}        & all  & 17{,}777 &   300 &  32 & Indoor rooms \\
    \bottomrule
  \end{tabular}
\end{table}

\section{Per-Category Analysis.}
\label{par:per-category_analysis}
\cref{tab:category} reports mAA@30° results broken down by the five failure categories.
Adding sparse correspondence prompts consistently improves performance across all categories.
The largest gains are observed for geometrically challenging cases such as large viewpoint changes (+12.1) and large-baseline pairs (+8.1), where automatic matching is particularly unreliable.
Substantial improvements are also seen for low-overlap pairs (+6.2) and scenes with extreme depth range (+6.6), suggesting that a few well-placed correspondences help stabilize geometric estimation.
Moderate viewpoint changes benefit as well (+4.3), though the baseline model already performs relatively well in this regime.
Overall, prompting improves the aggregate score from 54.6 to 61.7 (+7.1), demonstrating that sparse human-provided correspondences consistently help resolve challenging geometric configurations.

\begin{table}[t]                                                                                                                                                            
    \centering                                                                                                                                                                
    \caption{\textbf{Per-category results on the 2-view subset of Hard3DBench.} mAA@30\textdegree\,($\uparrow$) grouped by failure taxonomy.
      Categories are multi-label (pairs may belong to multiple categories).
      \mmethod{} with $K\!=\!8$ GT correspondences.
      {\tiny\color{teal}+x.x} denotes improvement over DUSt3R (retrained).}                                                                                                   
    \label{tab:category}                                                                                                                                                    
    \footnotesize                                                                                                                                                             
    \setlength{\tabcolsep}{3pt}                                                                                                                                             
    \begin{tabular}{@{}l cccccc@{}}
      \toprule
      Method
        & \scriptsize Lg.View.
        & \scriptsize Mod.View.
        & \scriptsize Low Ovlp.
        & \scriptsize Lg.Base.
        & \scriptsize Ext.Depth
        & \scriptsize All \\
      \midrule
      DUSt3R~\cite{dust3r}       & 35.9 & 52.0 & 46.3 & 44.6 & 38.9 & 41.8 \\
      \cdashline{1-7}\\[-8pt]
      DUSt3R (retrained)         & 54.3 & 56.3 & 48.7 & 58.0 & 41.3 & 54.6 \\
      \mmethod{} w/o prompts     & 53.2 & 56.3 & 46.5 & 56.6 & 39.4 & 53.1 \\
      \mmethod{} (8 prompts)     & \textbf{66.4}{\tiny\color{teal}+12.1} & \textbf{60.6}{\tiny\color{teal}+4.3} & \textbf{54.9}{\tiny\color{teal}+6.2} &
  \textbf{66.0}{\tiny\color{teal}+8.1} & \textbf{48.0}{\tiny\color{teal}+6.6} & \textbf{61.7}{\tiny\color{teal}+7.1} \\
      \bottomrule
    \end{tabular}
  \end{table}

\section{Additional Experiments}
\label{sec:supp_experiments}

\subsection{Ablation of Training Strategies on All Evaluation Sets}
\label{sec:supp_trainstrat}

\cref{tab:supp_trainstrat} extends the training strategy ablation of \cref{sec:experiments} to all five evaluation sets, with the same random GT clicks for every model.

\begin{table}[t]
  \centering
  \caption{\textbf{Ablation of training strategies on all evaluation sets} (mAA@30\textdegree, \%).
    All variants train the correspondence adapter. Same random GT clicks for every model. H3B: Hard3DBench 2-view. Best in \textbf{bold}.}
  \label{tab:supp_trainstrat}
  \small
  \setlength{\tabcolsep}{3.5pt}
  \begin{tabular}{@{}l cccccc@{}}
    \toprule
    Training & CO3D & ScanNet++ & ARKitScenes & H3B ID & H3B OOD & Mean \\
    \midrule
    \multicolumn{7}{@{}l}{\textit{No clicks}} \\[1pt]
    Adapter on frozen DUSt3R       & 79.2 & 82.6 & \textbf{86.0} & 39.2 & 47.1 & 66.8 \\
    Finetuned from pretrained DUSt3R & 83.4 & 88.2 & 79.8 & \textbf{55.5} & 49.7 & 71.3 \\
    Trained from scratch (ours)    & \textbf{84.4} & \textbf{89.9} & 84.3 & 55.3 & \textbf{50.8} & \textbf{72.9} \\
    \midrule
    \multicolumn{7}{@{}l}{\textit{One click}} \\[1pt]
    Adapter on frozen DUSt3R       & 84.0 & 90.1 & 84.8 & 49.7 & 47.2 & 71.2 \\
    Finetuned from pretrained DUSt3R & 83.7 & 88.9 & 80.3 & 56.7 & 51.6 & 72.3 \\
    Trained from scratch (ours)    & \textbf{85.0} & \textbf{91.1} & \textbf{85.2} & \textbf{57.5} & \textbf{56.6} & \textbf{75.1} \\
    \midrule
    \multicolumn{7}{@{}l}{\textit{Eight clicks}} \\[1pt]
    Adapter on frozen DUSt3R       & 85.3 & 91.0 & 85.2 & 52.4 & 50.4 & 72.9 \\
    Finetuned from pretrained DUSt3R & 83.8 & 89.3 & 80.3 & 57.3 & 54.4 & 73.0 \\
    Trained from scratch (ours)    & \textbf{85.9} & \textbf{92.2} & \textbf{87.3} & \textbf{61.8} & \textbf{62.4} & \textbf{77.9} \\
    \bottomrule
  \end{tabular}
\end{table}

\subsection{Stricter Pose Thresholds}
\label{sec:supp_strict}

\cref{tab:supp_maa_standard,tab:supp_maa_hard} extend \cref{tab:main,tab:hard} with mAA@5\textdegree{} and mAA@15\textdegree{}.
Against DUSt3R (retrained), the gain grows at stricter thresholds on the standard benchmark (grid oracle, $+15.7$ at 5\textdegree{} vs.\ $+5.7$ at 30\textdegree{}), where prompts sharpen reasonable poses, and at looser thresholds on Hard3DBench ($K\!=\!8$, $+3.6$ vs.\ $+7.7$), where prompts mainly rescue large failures.

\begin{table}[t]
    \centering
    \caption{\textbf{Pose accuracy at stricter thresholds on the standard benchmark.}
      mAA at 5\textdegree, 15\textdegree{} and 30\textdegree{} for every row of \cref{tab:main}, on the same pairs; the mAA@30\textdegree{} columns equal \cref{tab:main}.
      {\color{gray}Gray rows} are reference only and not ranked.
      {\tiny\color{teal}$\pm$x.x} denotes the difference from DUSt3R (retrained).
      Best in \textbf{bold}, second best \underline{underlined}.}
    \label{tab:supp_maa_standard}
    \small
    \setlength{\tabcolsep}{2pt}
    \resizebox{\linewidth}{!}{%
    \begin{tabular}{@{}l ccc ccc ccc@{}}
      \toprule
      & \multicolumn{3}{c}{CO3D} & \multicolumn{3}{c}{ScanNet++} & \multicolumn{3}{c}{ARKitScenes} \\
      \cmidrule(lr){2-4} \cmidrule(lr){5-7} \cmidrule(lr){8-10}
      Method & @5\textdegree & @15\textdegree & @30\textdegree & @5\textdegree & @15\textdegree & @30\textdegree & @5\textdegree & @15\textdegree & @30\textdegree \\
      \midrule
      {\color{gray} VGGT~\cite{vggt} (518)} & {\color{gray} 55.67} & {\color{gray} 79.22} & {\color{gray} 87.76} & {\color{gray} 21.80} & {\color{gray} 44.03} & {\color{gray} 59.47} & {\color{gray} 34.60} & {\color{gray} 63.85} & {\color{gray} 77.91} \\
      {\color{gray} Pi3X~\cite{pi3} (518)} & {\color{gray} 54.17} & {\color{gray} 79.26} & {\color{gray} 87.95} & {\color{gray} 47.80} & {\color{gray} 69.68} & {\color{gray} 80.77} & {\color{gray} 40.64} & {\color{gray} 69.55} & {\color{gray} 82.16} \\
      {\color{gray} VGGT~\cite{vggt} (224)} & {\color{gray} 4.37} & {\color{gray} 36.78} & {\color{gray} 62.53} & {\color{gray} 0.20} & {\color{gray} 7.14} & {\color{gray} 24.47} & {\color{gray} 1.26} & {\color{gray} 17.37} & {\color{gray} 40.89} \\
      {\color{gray} Pi3X~\cite{pi3} (224)} & {\color{gray} 44.27} & {\color{gray} 73.85} & {\color{gray} 85.02} & {\color{gray} 31.80} & {\color{gray} 57.70} & {\color{gray} 72.53} & {\color{gray} 31.68} & {\color{gray} 61.41} & {\color{gray} 76.39} \\[2pt]
      DUSt3R~\cite{dust3r}                 & 33.25 & 65.17 & 79.20 & 50.24 & 73.83 & 82.57 & \underline{48.46} & \underline{75.15} & \underline{86.02} \\
      MASt3R~\cite{mast3r}                 & 31.38 & 62.86 & 78.45 & 7.12 & 29.93 & 51.79 & 17.70 & 47.11 & 65.40 \\
      \cdashline{1-10}\\[-8pt]
      DUSt3R (retrained)                   & 45.63 & 73.53 & 84.33 & 60.02 & 83.11 & 90.81 & 42.48 & 70.23 & 82.71 \\
      \mmethod{} w/o prompts               & 46.12{\tiny\color{teal}+0.5} & 73.43{\tiny\color{teal}$-$0.1} & 84.41{\tiny\color{teal}+0.1} & 59.64{\tiny\color{teal}$-$0.4} & 82.05{\tiny\color{teal}$-$1.1} & 89.87{\tiny\color{teal}$-$0.9} & 46.18{\tiny\color{teal}+3.7} & 73.03{\tiny\color{teal}+2.8} & 84.27{\tiny\color{teal}+1.6} \\
      \mmethod{} (1p\_best5)               & \underline{47.20}{\tiny\color{teal}+1.6} & \underline{74.92}{\tiny\color{teal}+1.4} & \underline{85.71}{\tiny\color{teal}+1.4} & \underline{62.88}{\tiny\color{teal}+2.9} & \underline{84.43}{\tiny\color{teal}+1.3} & \underline{91.75}{\tiny\color{teal}+0.9} & 48.24{\tiny\color{teal}+5.8} & 75.01{\tiny\color{teal}+4.8} & 85.90{\tiny\color{teal}+3.2} \\
      \mmethod{} (grid oracle)             & \textbf{53.62}{\tiny\color{teal}+8.0} & \textbf{78.79}{\tiny\color{teal}+5.3} & \textbf{88.21}{\tiny\color{teal}+3.9} & \textbf{72.10}{\tiny\color{teal}+12.1} & \textbf{88.57}{\tiny\color{teal}+5.5} & \textbf{93.94}{\tiny\color{teal}+3.1} & \textbf{69.66}{\tiny\color{teal}+27.2} & \textbf{86.49}{\tiny\color{teal}+16.3} & \textbf{92.75}{\tiny\color{teal}+10.0} \\
      \bottomrule
    \end{tabular}}
  \end{table}

\begin{table}[t]
    \centering
    \caption{\textbf{Pose accuracy at stricter thresholds on Hard3DBench.}
      mAA at 5\textdegree, 15\textdegree{} and 30\textdegree{} for every row of \cref{tab:hard}, over finite poses; 5-view values pool the ten pairwise poses of each tuple. The mAA@30\textdegree{} columns equal \cref{tab:hard}.
      {\color{gray}Gray rows} are reference only and not ranked.
      {\tiny\color{teal}$\pm$x.x} denotes the difference from DUSt3R (retrained).
      Best in \textbf{bold}, second best \underline{underlined}.}
    \label{tab:supp_maa_hard}
    \small
    \setlength{\tabcolsep}{4pt}
    \begin{tabular}{@{}l ccc ccc@{}}
      \toprule
      & \multicolumn{3}{c}{\textit{2-view (pair)}} & \multicolumn{3}{c}{\textit{5-view (tuple)}} \\
      \cmidrule(lr){2-4} \cmidrule(lr){5-7}
      Method & @5\textdegree & @15\textdegree & @30\textdegree & @5\textdegree & @15\textdegree & @30\textdegree \\
      \midrule
      \multicolumn{7}{@{}l}{\textit{In-Distribution (404 pairs / 400 tuples)}} \\[2pt]
      {\color{gray} VGGT~\cite{vggt} (518)} & {\color{gray} 27.4} & {\color{gray} 50.1} & {\color{gray} 63.7} & {\color{gray} 32.5} & {\color{gray} 48.5} & {\color{gray} 56.5} \\
      {\color{gray} Pi3~\cite{pi3} (518)}  & {\color{gray} 34.3} & {\color{gray} 53.1} & {\color{gray} 64.5} & {\color{gray} 41.0} & {\color{gray} 54.4} & {\color{gray} 60.3} \\
      {\color{gray} VGGT~\cite{vggt} (224)} & {\color{gray} 3.5} & {\color{gray} 19.1} & {\color{gray} 40.5} & {\color{gray} 2.0} & {\color{gray} 16.9} & {\color{gray} 33.2} \\
      {\color{gray} Pi3~\cite{pi3} (224)}  & {\color{gray} 22.9} & {\color{gray} 46.7} & {\color{gray} 60.7} & {\color{gray} 30.1} & {\color{gray} 48.7} & {\color{gray} 56.9} \\[2pt]
      DUSt3R~\cite{dust3r}                 & 2.3 & 18.3 & 39.2 & 5.7 & 14.5 & 21.1 \\
      MASt3R~\cite{mast3r} (512)           & \textbf{21.8} & \textbf{46.1} & \textbf{62.9} & \underline{19.4} & 38.4 & 48.4 \\
      \cdashline{1-7}\\[-8pt]
      DUSt3R (retrained)                   & 15.3 & 38.4 & 56.1 & 15.6 & 30.6 & 39.4 \\
      \mmethod{} w/o prompts               & 15.7 & 37.9 & 55.3 & 15.8 & 30.6 & 38.8 \\
      \mmethod{} (1 prompt)                & 16.5{\tiny\color{teal}+1.2} & 39.6{\tiny\color{teal}+1.1} & 57.5{\tiny\color{teal}+1.3} & 17.6{\tiny\color{teal}+2.0} & 35.4{\tiny\color{teal}+4.8} & 45.0{\tiny\color{teal}+5.6} \\
      \mmethod{} (4 prompts)               & \underline{19.0}{\tiny\color{teal}+3.7} & 43.9{\tiny\color{teal}+5.5} & 61.3{\tiny\color{teal}+5.2} & \textbf{19.9}{\tiny\color{teal}+4.3} & \underline{38.6}{\tiny\color{teal}+8.0} & \underline{48.6}{\tiny\color{teal}+9.2} \\
      \mmethod{} (8 prompts)               & \underline{19.0}{\tiny\color{teal}+3.6} & \underline{44.1}{\tiny\color{teal}+5.7} & \underline{61.8}{\tiny\color{teal}+5.7} & \textbf{19.9}{\tiny\color{teal}+4.3} & \textbf{39.1}{\tiny\color{teal}+8.5} & \textbf{49.3}{\tiny\color{teal}+9.9} \\
      \midrule
      \multicolumn{7}{@{}l}{\textit{Out-of-Distribution (300 pairs / 300 tuples)}} \\[2pt]
      {\color{gray} VGGT~\cite{vggt} (518)} & {\color{gray} 52.1} & {\color{gray} 72.0} & {\color{gray} 81.3} & {\color{gray} 49.5} & {\color{gray} 70.5} & {\color{gray} 81.1} \\
      {\color{gray} Pi3~\cite{pi3} (518)}  & {\color{gray} 41.2} & {\color{gray} 68.9} & {\color{gray} 80.6} & {\color{gray} 44.6} & {\color{gray} 69.9} & {\color{gray} 81.4} \\
      {\color{gray} VGGT~\cite{vggt} (224)} & {\color{gray} 4.9} & {\color{gray} 26.0} & {\color{gray} 47.4} & {\color{gray} 4.4} & {\color{gray} 25.4} & {\color{gray} 47.1} \\
      {\color{gray} Pi3~\cite{pi3} (224)}  & {\color{gray} 24.1} & {\color{gray} 55.6} & {\color{gray} 71.1} & {\color{gray} 33.3} & {\color{gray} 60.2} & {\color{gray} 73.5} \\[2pt]
      DUSt3R~\cite{dust3r}                 & 9.1 & 27.3 & 45.3 & 15.9 & 40.2 & 54.7 \\
      MASt3R~\cite{mast3r} (512)           & \underline{14.4} & 33.2 & 51.6 & \textbf{19.6} & \underline{42.4} & 58.0 \\
      \cdashline{1-7}\\[-8pt]
      DUSt3R (retrained)                   & 10.3 & 36.2 & 52.6 & 14.4 & 39.0 & 54.8 \\
      \mmethod{} w/o prompts               & 11.1 & 36.0 & 50.8 & 13.9 & 38.9 & 54.7 \\
      \mmethod{} (1 prompt)                & 12.7{\tiny\color{teal}+2.3} & 40.3{\tiny\color{teal}+4.0} & 56.6{\tiny\color{teal}+4.0} & 14.2{\tiny\color{teal}$-$0.2} & 40.0{\tiny\color{teal}+1.0} & 56.7{\tiny\color{teal}+1.9} \\
      \mmethod{} (4 prompts)               & 14.2{\tiny\color{teal}+3.9} & \underline{44.1}{\tiny\color{teal}+7.8} & \underline{60.4}{\tiny\color{teal}+7.8} & 15.2{\tiny\color{teal}+0.8} & 41.7{\tiny\color{teal}+2.8} & \underline{58.6}{\tiny\color{teal}+3.8} \\
      \mmethod{} (8 prompts)               & \textbf{15.3}{\tiny\color{teal}+5.0} & \textbf{45.2}{\tiny\color{teal}+8.9} & \textbf{62.4}{\tiny\color{teal}+9.8} & \underline{16.0}{\tiny\color{teal}+1.6} & \textbf{43.1}{\tiny\color{teal}+4.1} & \textbf{60.0}{\tiny\color{teal}+5.2} \\
      \bottomrule
    \end{tabular}
  \end{table}

\subsection{Effect of the Number of Prompts}
\label{sec:supp_k_ablation}

\begin{figure}[t]
  \centering
  \includegraphics[width=0.75\linewidth]{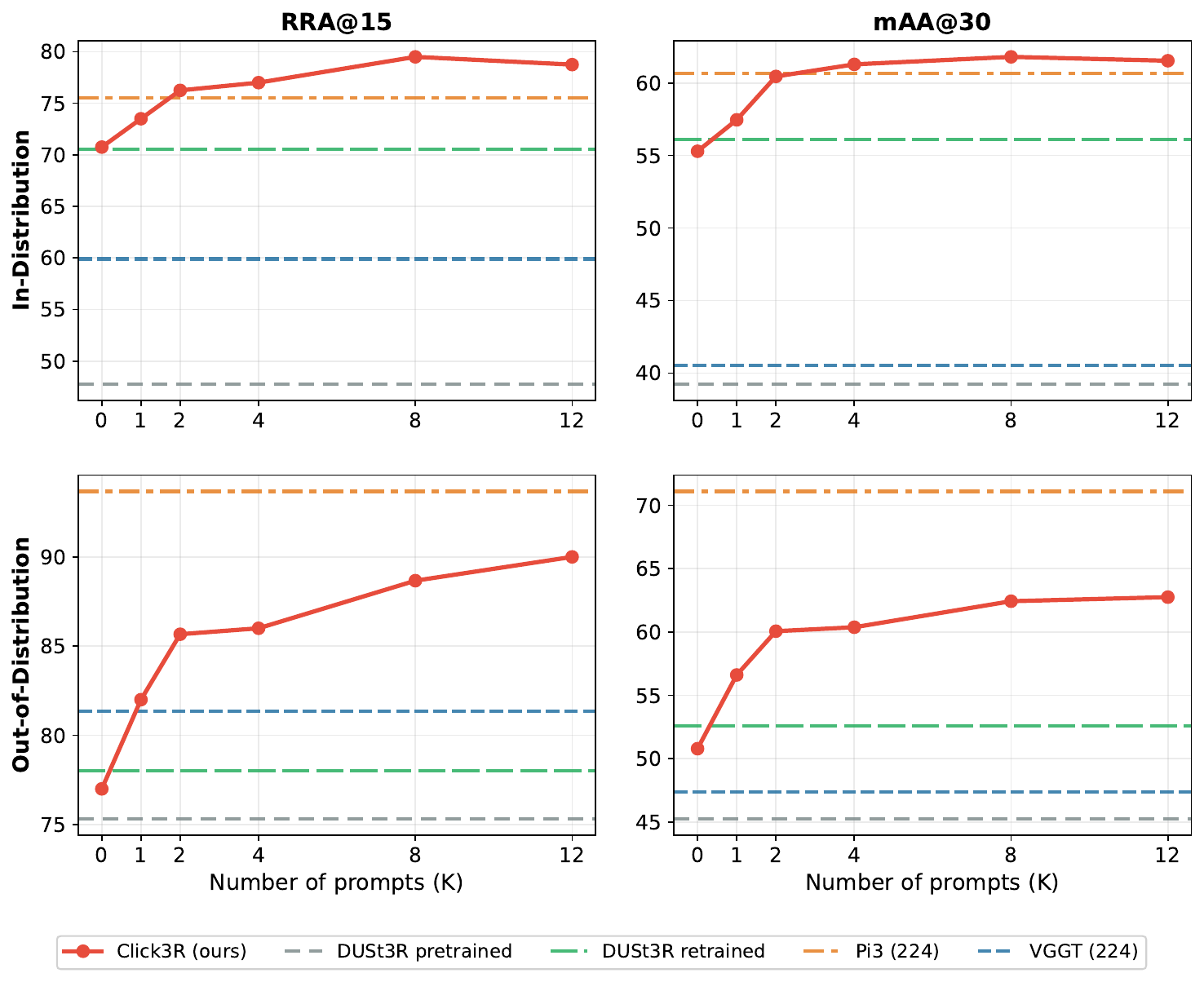}
  \caption{\textbf{Prompt count ablation on Hard3DBench.}
    Pose accuracy of \mmethod{} as a function of $K$ (red), with baselines shown as horizontal lines.
    Top: in-distribution (404 pairs); bottom: out-of-distribution (300 pairs).
    All methods use $224\!\times\!224$ resolution.}
  \label{fig:k_ablation}
\end{figure}

\cref{fig:k_ablation} shows pose accuracy as a function of the number of correspondence prompts $K$ on Hard3DBench.
We vary $K \in \{0, 1, 2, 4, 8, 12\}$, where $K\!=\!0$ denotes the unprompted mode in which the adapter is bypassed entirely.
All methods use $224\!\times\!224$ input for a fair comparison.

Accuracy generally improves with $K$ on both splits.
The effect is more pronounced on the out-of-distribution split, where even a single click ($K\!=\!1$) yields a substantial gain.
The improvement is larger on the out-of-distribution split ($+11.7$ pp RRA@15 from $K\!=\!0$ to $K\!=\!8$) than in-distribution ($+8.8$ pp), consistent with the base model having more room for improvement on unseen data.
With $K\!\geq\!2$, \mmethod{} outperforms both DUSt3R variants and VGGT on both RRA@15 and mAA@30.
On the OOD split, Pi3~\cite{pi3} remains stronger at every $K$, likely due to its larger training data and model capacity.

\subsection{Robustness to Image Augmentations}
\label{sec:supp_augmentation}

We test how well \mmethod{} tolerates photometric perturbations by augmenting only view~2 while keeping view~1 unchanged, and measuring RRA@15° on 100 hard OOD pairs.
We consider two augmentation types at two severity levels each: brightness scaling ($\times 0.5$, $\times 1.5$) and additive Gaussian noise ($\sigma \in \{0.1, 0.3\}$).

\begin{figure}[t]
  \centering
  \includegraphics[width=0.68\linewidth]{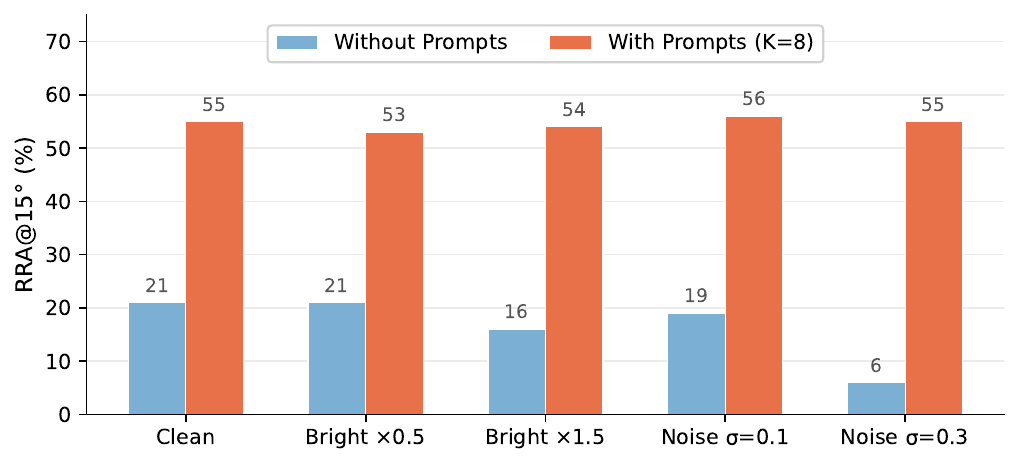}
  \caption{\textbf{Robustness to photometric augmentations.}
    We perturb view~2 with brightness scaling or Gaussian noise and report RRA@15° on 100 hard OOD pairs.
    Prompts consistently improve accuracy, with the largest margin under heavy noise ($\sigma{=}0.3$): 55\% vs.\ 6\%.}
  \label{fig:augmentation_robustness}
\end{figure}

As shown in \cref{fig:augmentation_robustness}, brightness changes have negligible impact on both the unprompted and prompted models.
Gaussian noise, however, exposes a clear robustness gap.
At $\sigma{=}0.3$ the unprompted model drops from 21\% to 6\% RRA@15°, whereas the prompted model stays at 55\%, suggesting that correspondence prompts provide geometric constraints that compensate for corrupted visual features.

\subsection{Robustness to Incorrect Prompts}
\label{sec:supp_wrong}

\cref{tab:supp_wrong} complements the perturbation analysis of Section~5.5 with completely incorrect clicks: pose accuracy degrades monotonically as incorrect clicks replace correct ones, and all-incorrect clicks fall below using no prompts.

\begin{table}[t]
  \centering
  \caption{\textbf{Robustness to incorrect prompts.}
    On the 100 pairs of \cref{tab:human}, we use 8 random ground-truth clicks and replace a fraction of the view-2 clicks by locations sampled uniformly over the image, at least 16 pixels from the true correspondence. Results are pooled over 5 random draws per pair.}
  \label{tab:supp_wrong}
  \small
  \begin{tabular}{@{}lcc@{}}
    \toprule
    Correct clicks (of 8) & medRRE\,$\downarrow$ & mAA@30\,$\uparrow$ \\
    \midrule
    100\% & 7.70\textdegree & 62.8 \\
    75\% & 8.81\textdegree & 55.6 \\
    50\% & 12.08\textdegree & 44.9 \\
    25\% & 20.63\textdegree & 30.0 \\
    0\% & 34.76\textdegree & 17.7 \\
    \midrule
    No prompts & 18.64\textdegree & 35.6 \\
    \bottomrule
  \end{tabular}
\end{table}

\subsection{Automatic Prompts}
\label{sec:supp_auto}

We also consider scenarios without ground-truth correspondences, where we use automatic prompting from RoMa v2~\cite{edstedt2026roma}, a state-of-the-art dense feature matcher. We run it on each image pair, rank its matches by confidence, and greedily select $K$ of them with a spacing of 10\% of the image width in both views, so that the clicks spread over the scene. These clicks improve mAA@30 on all three sets of \cref{tab:supp_auto}: with $K\!=\!4$, from 55.9 to 58.0 on the in-distribution pairs, from 49.4 to 59.6 on the out-of-distribution pairs, and from 35.6 to 55.0 on the 100 out-of-distribution pairs with human clicks, close to the human clicks themselves (56.8).

\begin{table}[t]
  \centering
  \caption{\textbf{Automatic prompts from RoMa v2~\cite{edstedt2026roma}.}
    mAA@30\textdegree{} of \mmethod{} with clicks from RoMa v2 matches or from random ground-truth correspondences (GT). H3B: Hard3DBench 2-view (404 ID, 300 OOD pairs); the subset is the 100 OOD pairs with human clicks.}
  \label{tab:supp_auto}
  \small
  \setlength{\tabcolsep}{5pt}
  \begin{tabular}{@{}lcc@{\quad}cc@{\quad}cc@{}}
    \toprule
     & \multicolumn{2}{c}{H3B ID} & \multicolumn{2}{c}{H3B OOD} & \multicolumn{2}{c}{H3B OOD subset} \\
    \cmidrule(lr){2-3}\cmidrule(lr){4-5}\cmidrule(l){6-7}
    Clicks & RoMa v2 & GT & RoMa v2 & GT & RoMa v2 & GT \\
    \midrule
    $K\!=\!0$ & \multicolumn{2}{c}{55.9} & \multicolumn{2}{c}{49.4} & \multicolumn{2}{c}{35.6} \\
    \midrule
    $K\!=\!1$ & 58.4 & 57.8 & 57.0 & 55.8 & 48.7 & 48.1 \\
    $K\!=\!4$ & 58.0 & 60.5 & 59.6 & 60.8 & 55.0 & 58.7 \\
    $K\!=\!8$ & 58.4 & 62.4 & 59.4 & 61.8 & 58.5 & 61.1 \\
    \midrule
    Human & \multicolumn{2}{c}{--} & \multicolumn{2}{c}{--} & \multicolumn{2}{c}{56.8} \\
    \bottomrule
  \end{tabular}
\end{table}

\subsection{Transfer to MASt3R}
\label{sec:supp_mast3r}

As a controlled study of whether correspondence prompting transfers to another backbone, we add the adapter to MASt3R~\cite{mast3r} and train it end-to-end, along with a control without the adapter on the same data. Both are trained on the same nine datasets as \mmethod{} (CO3D, ARKitScenes, ScanNet++, WildRGB-D, BlendedMVS, MegaDepth, Habitat, Waymo, and StaticThings3D), a smaller mixture than the official MASt3R recipe, so their absolute numbers are not comparable with the released model. Test-time clicks raise mAA@30 from 61.1 to 62.4 on the in-distribution pairs and from 52.1 to 54.9 on the out-of-distribution pairs (\cref{tab:supp_mast3r}), confirming that correspondence clicks also improve performance on a different backbone.

\begin{table}[t]
  \centering
  \caption{\textbf{Correspondence prompting on MASt3R.}
    mAA@30\textdegree{} on Hard3DBench 2-view at 512 resolution (404 ID, 300 OOD pairs). $K\!=\!8$ uses random ground-truth clicks.}
  \label{tab:supp_mast3r}
  \small
  \begin{tabular}{@{}lcc@{}}
    \toprule
     & H3B ID & H3B OOD \\
    \midrule
    MASt3R (retrained) & 61.3 & 50.6 \\
    MASt3R + adapter, $K\!=\!0$ & 61.1 & 52.1 \\
    MASt3R + adapter, $K\!=\!8$ & 62.4 & 54.9 \\
    \bottomrule
  \end{tabular}
\end{table}

\subsection{Error Distribution Analysis}
\label{sec:supp_distribution}

We plot per-sample distributions of reconstruction loss and pose error across the three evaluation datasets of \cref{tab:main}, comparing \mmethod{} against DUSt3R~\cite{dust3r}, MASt3R~\cite{mast3r}, VGGT~\cite{vggt}, and Pi3X~\cite{pi3} at $224\!\times\!224$ resolution.

\begin{figure}[t]
  \centering
  \includegraphics[width=0.68\linewidth]{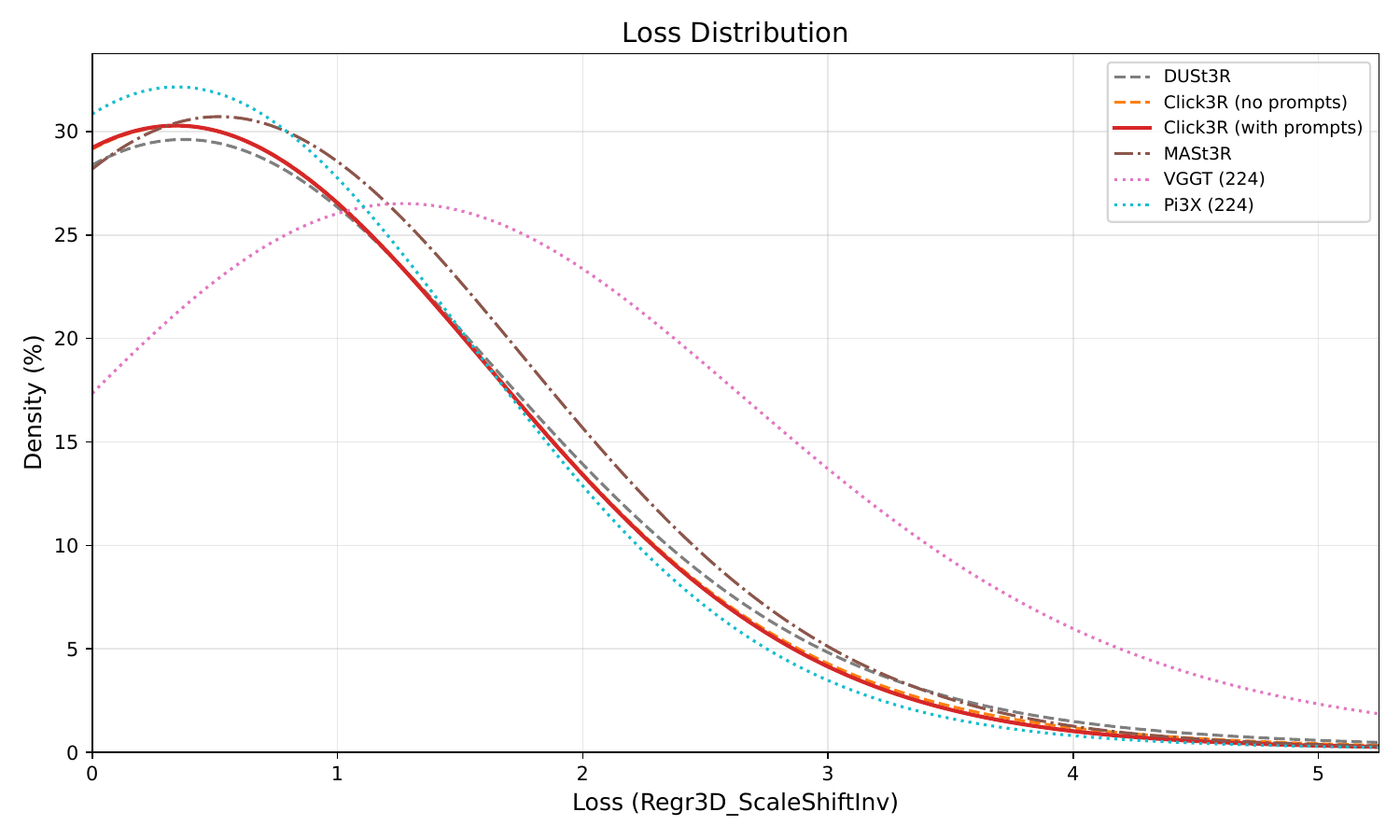}
  \caption{\textbf{Reconstruction loss distribution} (aggregated over the three datasets of \cref{tab:main}).
    Density estimated via KDE with Scott's bandwidth.
    \mmethod{} concentrates more mass near zero and exhibits a shorter tail than DUSt3R.}
  \label{fig:loss_dist}
\end{figure}

\paragraph{Reconstruction loss.}
\cref{fig:loss_dist} shows the distribution of the scale-and-shift-invariant pointmap loss.
Both \mmethod{} variants (with and without prompts) have more mass near zero than DUSt3R:
the fraction of samples with loss $> 1.0$ drops from 14.9\% (DUSt3R) to 11.4\% (\mmethod{}), and with loss $> 2.0$ from 6.1\% to 3.8\%.
\mmethod{} without prompts already accounts for most of this reduction.
VGGT~\cite{vggt} at $224\!\times\!224$ shows a broader distribution (63.0\% of samples exceeding loss $1.0$), reflecting its sensitivity to input resolution.

\begin{figure}[t]
  \centering
  \includegraphics[width=0.68\linewidth]{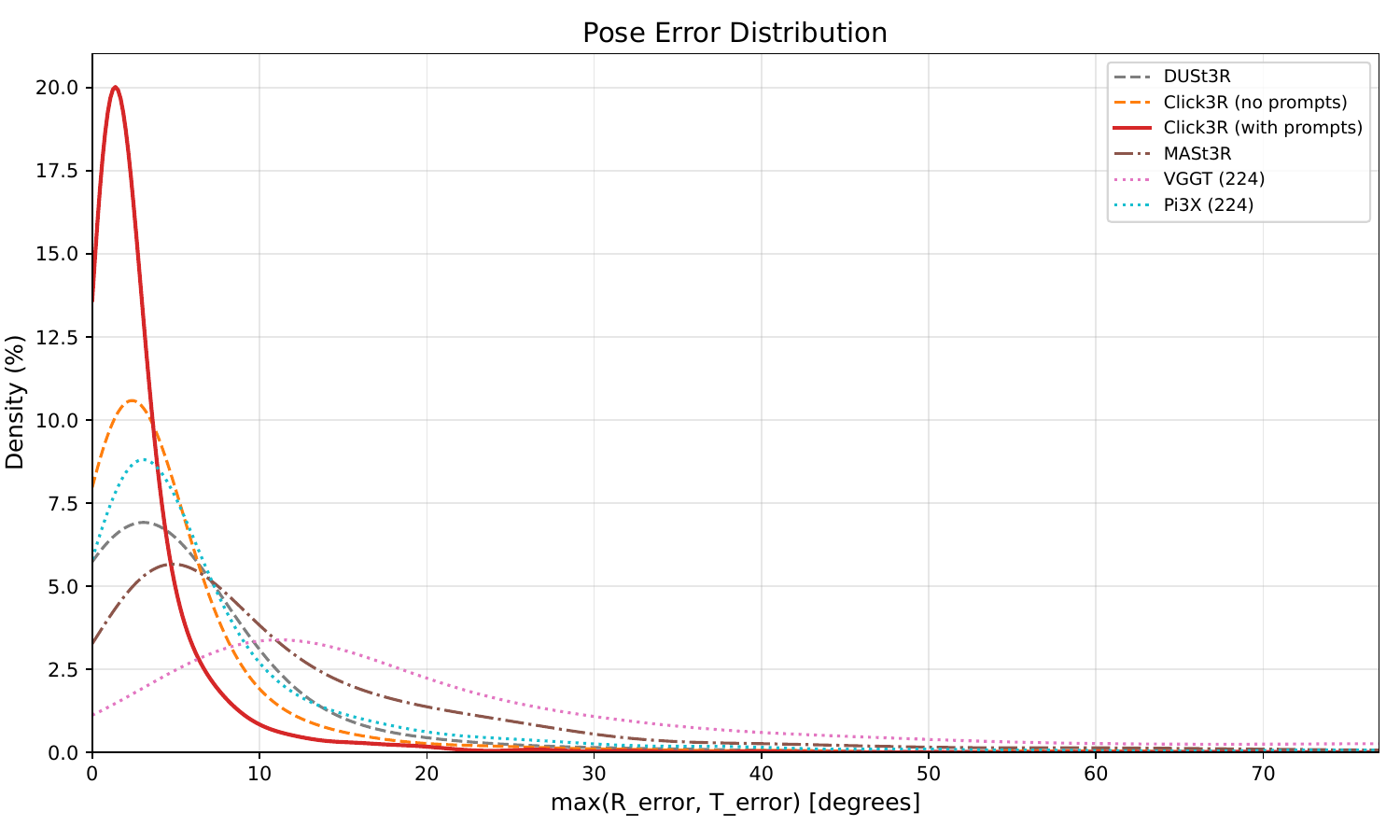}
  \caption{\textbf{Pose error distribution} (aggregated over the three datasets of \cref{tab:main} at $224\!\times\!224$, $\max(R_\text{err}, T_\text{err})$ in degrees).
    \mmethod{} with prompts produces the sharpest peak near $0^\circ$ and the shortest tail.
    The fraction of pairs exceeding $15^\circ$ error drops from 9.4\% (DUSt3R) to 2.8\% (\mmethod{} with prompts), a $3.4{\times}$ reduction in pose failures.}
  \label{fig:pose_dist}
\end{figure}

\paragraph{Pose error.}
\cref{fig:pose_dist} shows the distribution of the combined pose error $\max(R_\text{err}, T_\text{err})$.
\mmethod{} with prompts produces a sharper peak near $0^\circ$ and a shorter tail than all baselines.
The fraction of pairs with pose error exceeding $15^\circ$ drops from 9.4\% (DUSt3R) to 6.1\% (\mmethod{} without prompts) and further to 2.8\% (\mmethod{} with prompts)---a $3.4{\times}$ reduction.
At the $30^\circ$ threshold, the failure rate decreases from 4.7\% to 0.8\%.
The reduction is concentrated in the tail rather than a uniform shift of the distribution, indicating that prompts primarily resolve the hardest cases.

\section{What Does the Adapter Do?}
\label{sec:supp_adapter}

We examine how the correspondence adapter modifies encoder features and how this affects downstream decoder attention.

\begin{figure}[t]
  \centering
  \includegraphics[width=0.9\linewidth]{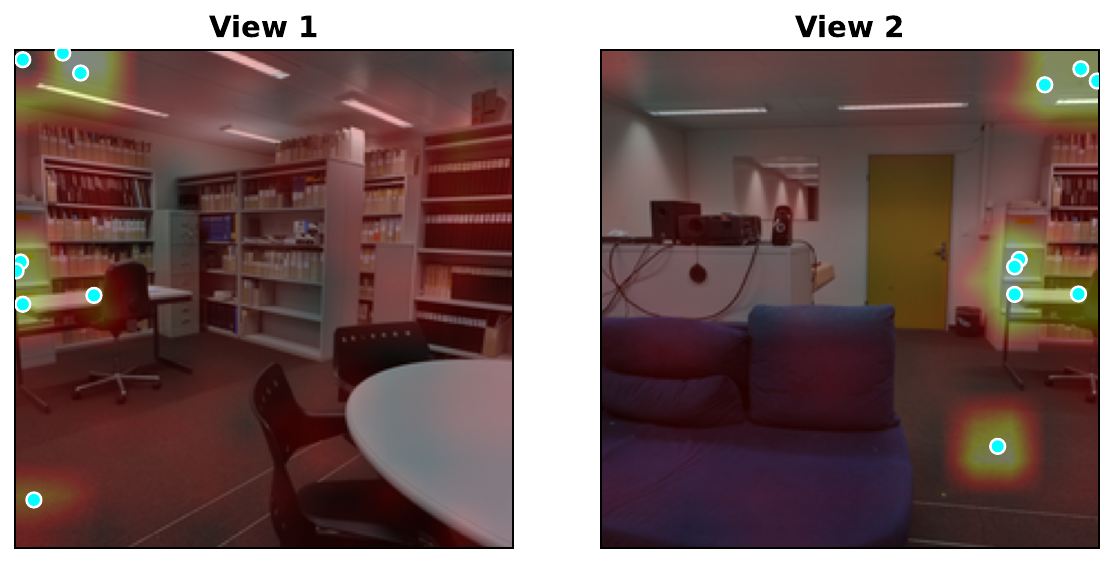}
  \caption{\textbf{What the adapter changes.}
    Per-token relative feature change induced by the adapter, with $K\!=\!8$ prompt locations shown as cyan dots.
    Bright regions indicate tokens whose features were modified most (each view uses its own color scale).
    The adapter updates every token, most strongly near the prompt correspondences.}
  \label{fig:adapter_attention}
\end{figure}

\paragraph{Feature modification.}
\cref{fig:adapter_attention} visualizes the relative change in each token's feature vector before and after the adapter ($\lVert\Delta f\rVert / \lVert f\rVert$) for a representative pair with $K\!=\!8$ prompts.
The adapter does not apply a uniform transformation: feature changes are largest near the prompt locations (cyan dots), although every token is modified.

\begin{figure}[t]
  \centering
  \includegraphics[width=\linewidth]{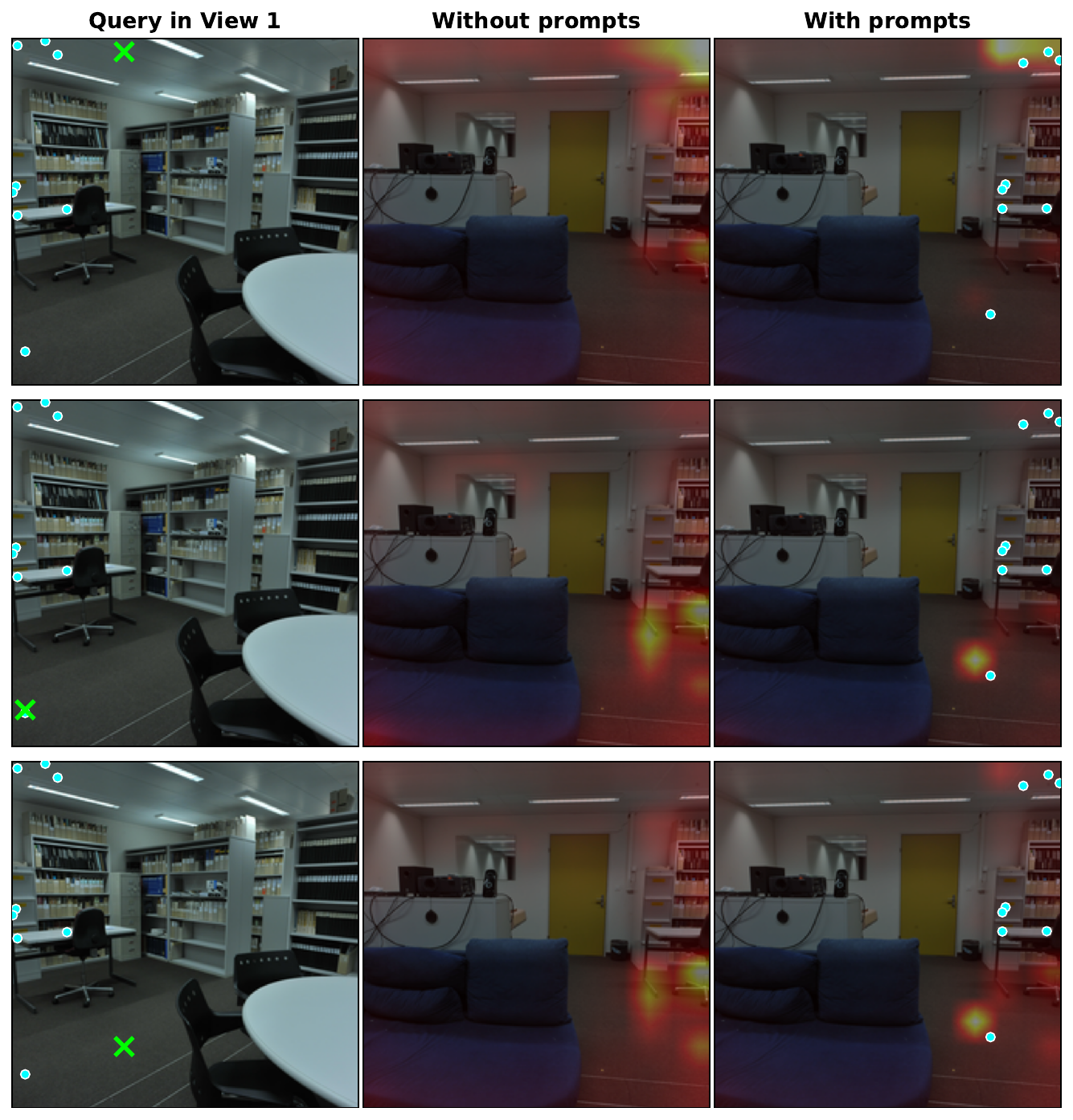}
  \caption{\textbf{How prompts reshape decoder cross-attention.}
    Each row shows a query patch in view~1 (green cross) and its decoder cross-attention map over view~2.
    Without prompts (middle), attention is diffuse.
    With $K\!=\!8$ prompts (right, cyan dots), attention concentrates on the correct corresponding region.}
  \label{fig:cross_attn_comparison}
\end{figure}

\paragraph{Effect on decoder cross-attention.}
\cref{fig:cross_attn_comparison} shows how prompts reshape the decoder's cross-view attention.
For three query patches in view~1 (green crosses), we visualize where each query attends in view~2 with and without prompts.
Without prompts, cross-attention is diffuse and sometimes focuses on incorrect regions.
With prompts, attention shifts toward the geometrically correct correspondences.

\begin{figure}[t]
  \centering
  \includegraphics[width=0.95\linewidth]{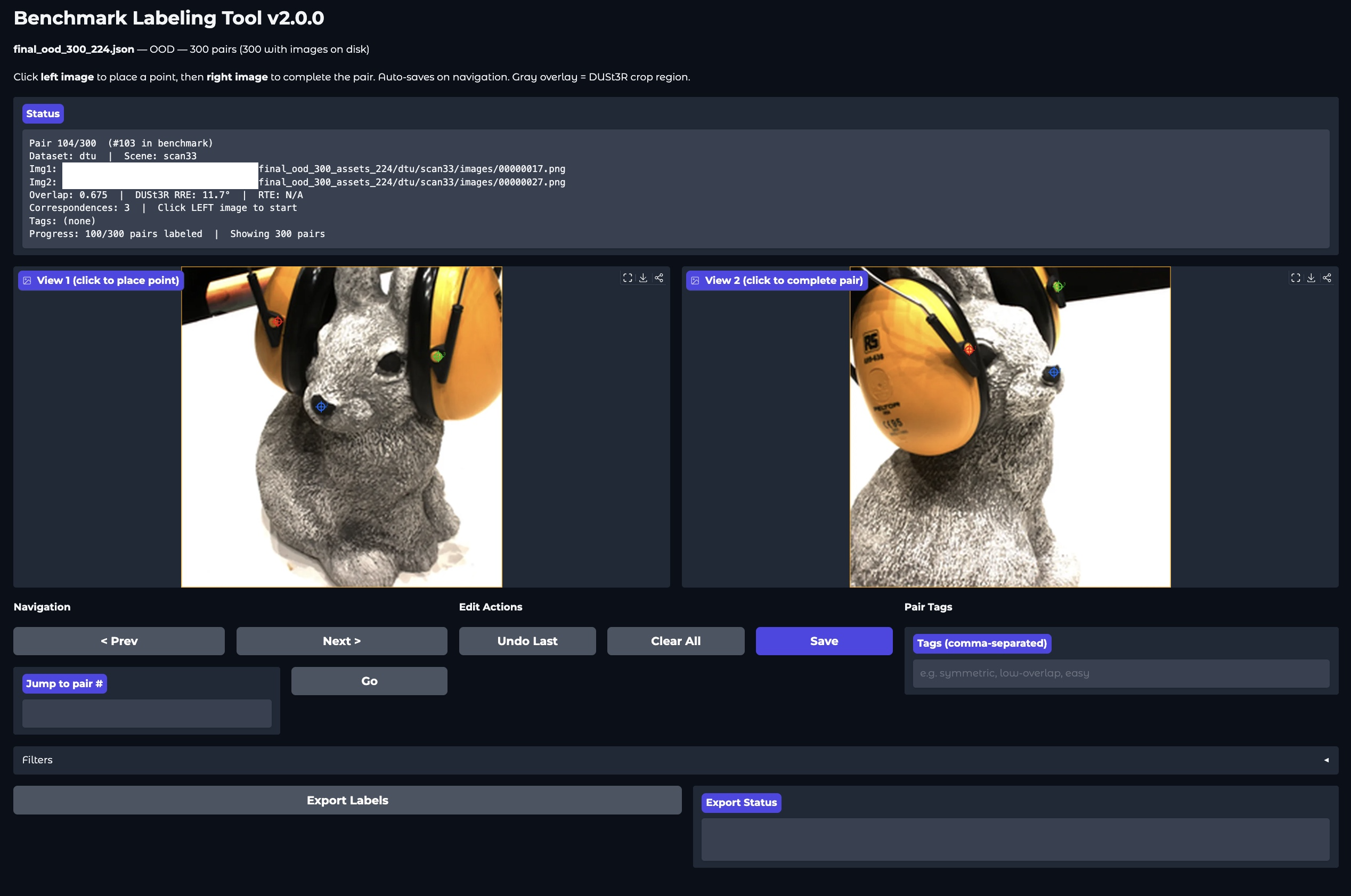}
  \caption{A screenshot of the \textbf{labeling tool} we build to harness human labeled correspondences.}
  \label{fig:labeling_tool}
\end{figure}

\section{More Qualitative Visualizations}
\label{sec:supp_qualitative}

\begin{figure}[t]
  \centering
  \includegraphics[width=0.95\linewidth]{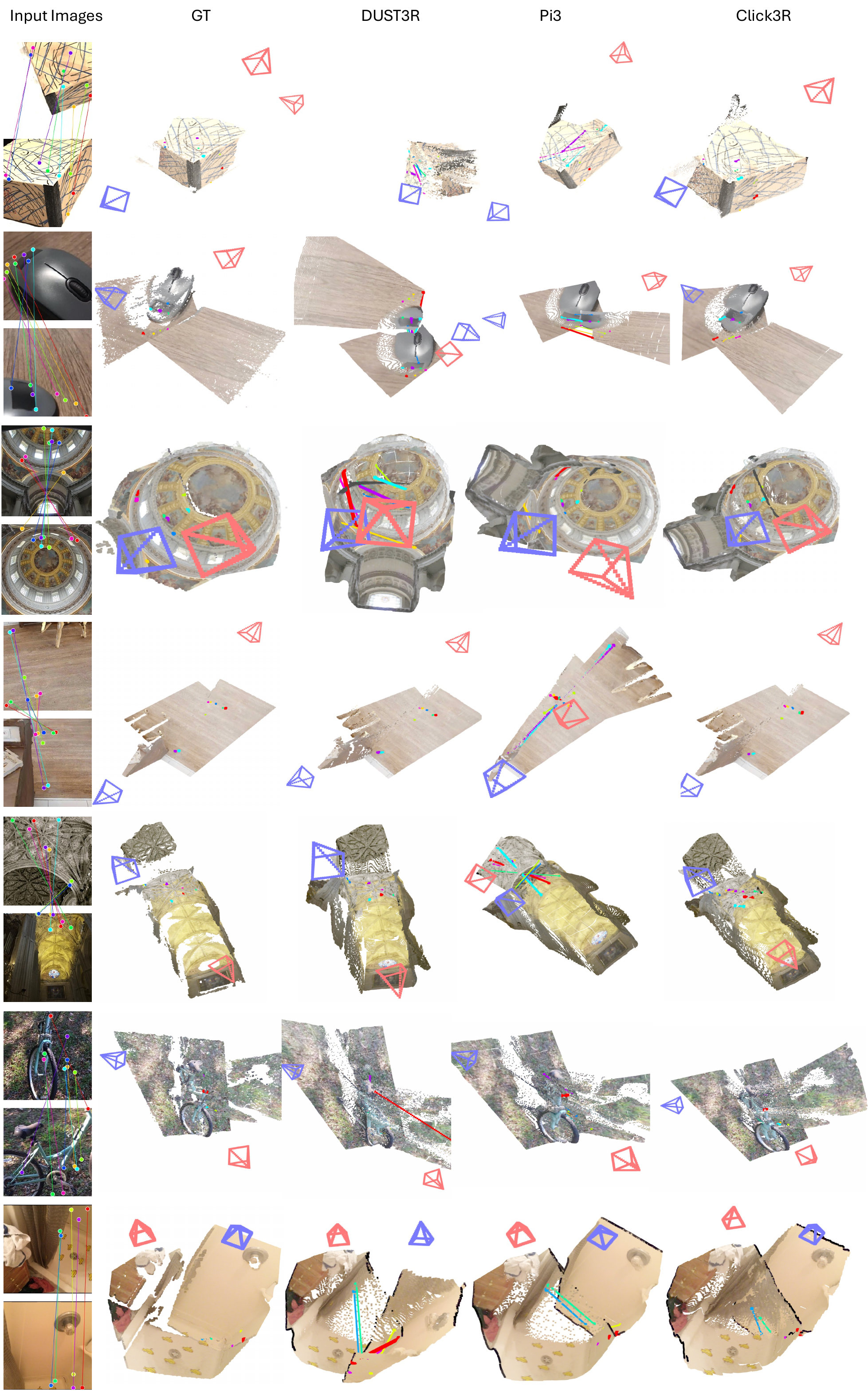}
  \caption{\textbf{Additional qualitative comparisons.} On challenging pairs, different baselines fail in different ways: DUSt3R may produce a distorted reconstruction on a pair where Pi3 succeeds, or vice versa. \mmethod{} with sparse correspondence clicks consistently recovers accurate geometry.}
  \label{fig:supp_qualitative}
\end{figure}

\cref{fig:supp_qualitative} presents additional qualitative comparisons on challenging pairs.
For each example, we show the input pair with correspondence clicks, the ground-truth reconstruction, and the predictions of DUSt3R, Pi3, and \mmethod{}.
These cases illustrate scenarios where at least one baseline produces a noticeably incorrect reconstruction. For instance, DUSt3R may fail on a pair where Pi3 succeeds, or vice versa.
\mmethod{} with sparse correspondence clicks consistently recovers geometry close to the ground truth across all shown examples.
The clicked correspondences converge to consistent 3D locations in the \mmethod{} reconstruction, confirming that the adapter successfully leverages sparse correspondences to resolve geometric ambiguities that affect different baselines in different ways.

\section{Failure Cases}
\label{sec:supp_failures}

\begin{figure}[t]
  \centering
  \begin{subfigure}{0.48\linewidth}
    \centering
    \includegraphics[width=\linewidth]{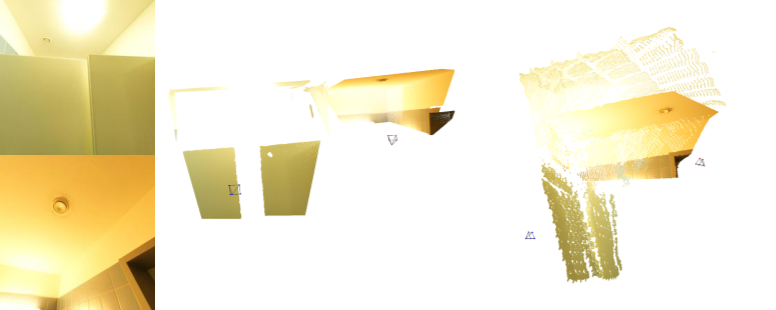}
    \caption{Few overlap}
    \label{fig:fail_few_overlap}
  \end{subfigure}
  \hfill
  \begin{subfigure}{0.48\linewidth}
    \centering
    \includegraphics[width=\linewidth]{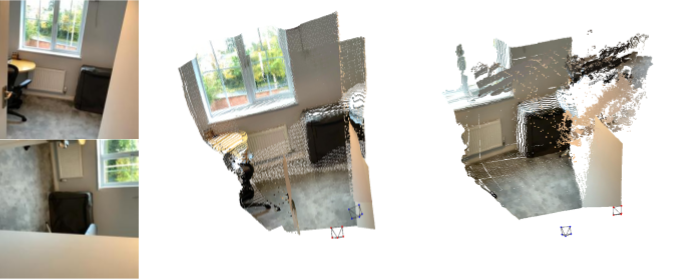}
    \caption{Heavy rotation}
    \label{fig:fail_heavy_rot}
  \end{subfigure}

  \vspace{0.5em}

  \begin{subfigure}{0.48\linewidth}
    \centering
    \includegraphics[width=\linewidth]{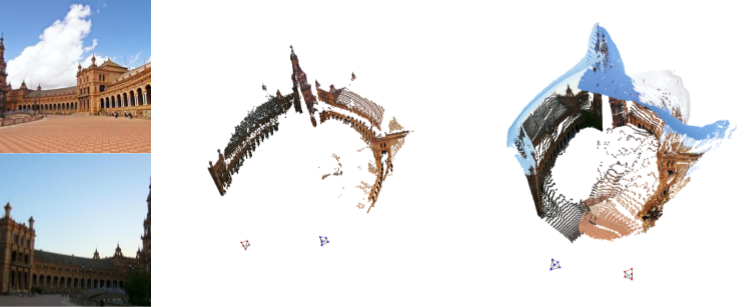}
    \caption{Large-scale scene}
    \label{fig:fail_large_scale}
  \end{subfigure}
  \hfill
  \begin{subfigure}{0.48\linewidth}
    \centering
    \includegraphics[width=\linewidth]{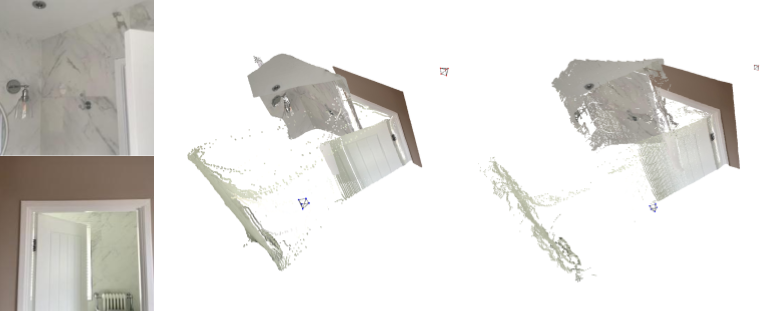}
    \caption{Ambiguous correspondences}
    \label{fig:fail_ambiguous}
  \end{subfigure}

  \caption{\textbf{Failure cases of \mmethod{}.}
    Each panel: input pair (left), ground truth (middle), \mmethod{} with clicks (right).
    (a)~Near-zero overlap showcasing fundamental limitation of click-based supervision.
    (b)~Extreme rotations cause geometric distortion that clicks cannot resolve.
    (c)~In large-scale scenes, regions far from clicked correspondences still deform.
    (d)~Clicks on featureless surfaces are inherently ambiguous and provide insufficient geometric signal.}
  \label{fig:failures}
\end{figure}

\cref{fig:failures} shows four representative failure modes of \mmethod{}.
Each panel shows the input image pair (left), the ground-truth reconstruction (middle), and the \mmethod{} reconstruction with clicks (right).

\textbf{(a)~Few overlap.}
When the two views share little or no visual overlap, the model must hallucinate the relative geometry from appearance alone.
In this ScanNet++ example the views observe disjoint parts of a room; comparing the ground truth with the \mmethod{} output reveals a severely misaligned reconstruction.
Without co-visible content, correspondence clicks cannot anchor the two views---this is a fundamental limitation of pairwise reconstruction.

\textbf{(b)~Heavy rotation.}
Extreme viewpoint changes ($>90^\circ$) remain an unsolved challenge.
In this ARKitScenes pair, the \mmethod{} reconstruction remains geometrically distorted compared to the ground truth.
While clicks can reduce error in moderate-rotation cases (\cref{tab:category}), very heavy rotations overwhelm the model's capacity to reconcile drastically different visual appearances.

\textbf{(c)~Large-scale scenes.}
For outdoor scenes with large spatial extent (MegaDepth), correspondence clicks correctly constrain geometry near the clicked regions, but distant structures still exhibit bending and deformation---visible here as warped sky and distorted building facades far from the click locations.
A small number of clicks cannot regularize the entire scene when the depth range spans tens of meters.

\textbf{(d)~Ambiguous correspondences.}
On textureless or repetitive surfaces, correspondences themselves become ambiguous.
In this ARKitScenes bathroom scene with featureless marble walls and white surfaces, a click on one location is visually indistinguishable from many others, providing little geometric information.
The \mmethod{} reconstruction remains distorted compared to the ground truth, and clicks offer negligible improvement.

\section{Licenses of Existing Assets}
\label{sec:supp_licenses}

We acknowledge and respect the licenses of all assets used in this work.
\cref{tab:supp_licenses} lists the license of each dataset, code base, and pretrained checkpoint we rely on.
All assets are used in compliance with their license terms, and our work is intended for non-commercial research purposes consistent with these licenses.

\begin{table}[t]
  \centering
  \caption{\textbf{Licenses of assets used.}}
  \label{tab:supp_licenses}
  \footnotesize
  \setlength{\tabcolsep}{4pt}
  \begin{tabular}{@{}ll@{}}
    \toprule
    Asset & License \\
    \midrule
    \multicolumn{2}{@{}l}{\textit{Datasets (training)}} \\[2pt]
    CO3D~\cite{co3d}                    & CC-BY-NC 4.0 \\
    ARKitScenes~\cite{arkitscenes}      & Apple License (non-commercial primary) \\
    ScanNet++~\cite{scannetpp}          & ScanNet++ Terms of Use \\
    WildRGB-D~\cite{wildrgbd}           & MIT License \\
    BlendedMVS~\cite{blendedmvs}        & CC-BY 4.0 \\
    MegaDepth~\cite{megadepth}          & CC-BY 4.0 (depth, SfM); images under their Flickr licenses \\
    Habitat-Sim~\cite{habitat}          & MIT License \\
    Waymo Open Dataset~\cite{xu2025wod} & Waymo Dataset License (non-commercial) \\
    \midrule
    \multicolumn{2}{@{}l}{\textit{Datasets (Hard3DBench evaluation)}} \\[2pt]
    7Scenes~\cite{7scenes}              & Microsoft Research License (non-commercial) \\
    DTU~\cite{dtu}                      & No formal license; cite the original paper \\
    ETH3D~\cite{eth3d}                  & CC-BY-NC-SA 4.0 \\
    NRGBD~\cite{nrgbd}                  & CC-BY / CC-0 (per synthetic scene) \\
    ScanNet~\cite{scannet}              & ScanNet Terms of Use (data); MIT License (code) \\
    \midrule
    \multicolumn{2}{@{}l}{\textit{Code / Models}} \\[2pt]
    DUSt3R~\cite{dust3r}                & CC-BY-NC-SA 4.0 \\
    CroCo v2~\cite{crocov2}             & CC-BY-NC-SA 4.0 \\
    Segment Anything (SAM)~\cite{sam}   & Apache License 2.0 \\
    MASt3R~\cite{mast3r}                & CC-BY-NC-SA 4.0 \\
    VGGT~\cite{vggt}                    & VGGT License (custom; weights non-commercial) \\
    Pi3~\cite{pi3}                      & BSD 3-Clause (code); CC-BY-NC 4.0 (weights) \\
    \bottomrule
  \end{tabular}
\end{table}

\section{Limitations}
\label{sec:limitations}
Although correspondence prompts are ideally injected into a frozen pretrained model, we find that training the base model~\cite{dust3r} end-to-end with the adapter is necessary in practice, incurring nontrivial training cost. Partly due to this cost, our experiments focus on two-view reconstruction; extending to multi-view models~\cite{pi3,vggt,fast3r,mvdust3rplus} with trajectory prompts across $N$ views is an important direction.

Architecturally, the adapter treats correspondences as soft feature conditioning via cross-attention rather than enforcing hard geometric constraints. Although eight correspondences suffice to determine relative pose algebraically~\cite{hartley2003multiple}, the adapter cannot replicate such reasoning: it modulates all image tokens through learned attention at a single injection point, and this signal is progressively diluted across twelve subsequent decoder cross-attention blocks. %
Deeper injection at multiple decoder layers and explicit geometric reasoning modules may be worth future inquiry to help close this gap.

Finally, due to computational constraints, the model is trained only at a resolution of $224$px in a stereo setting, although it has demonstrated multi-view capability through global alignment. In addition, prompts must be co-visible in both views, so the two inputs need a shared visible region; non-overlapping inputs could be connected through intermediate views by chaining correspondences. Severely incorrect correspondences can degrade reconstruction quality (\cref{tab:supp_wrong}).

\section{Broader Impacts}
\label{sec:supp_broader_impacts}

Our work enables more efficient 3D reconstruction from sparse user-provided correspondences. By allowing a small number of clicks to anchor pairwise reconstruction, \mmethod{} can lower the barrier to 3D capture in diverse, in-the-wild scenes, which may benefit downstream applications such as robotics, AR/VR content creation, and scientific imaging where dense ground-truth geometry is unavailable. Our model and benchmark are built from publicly available datasets (CO3D, ScanNet++, ARKitScenes, BlendedMVS, MegaDepth, Habitat, WildRGB-D, Waymo, and standard evaluation benchmarks), and therefore do not raise privacy concerns. While our algorithm itself is scene-agnostic and does not introduce explicit bias, the underlying training datasets may reflect societal or geographic biases present in the source data (e.g., predominantly indoor scenes from North American buildings, or object categories skewed toward consumer products). We encourage future work to investigate and mitigate such biases when deploying systems trained on these data.

\newpage
\clearpage

\end{document}